\documentclass[11pt]{article}

\usepackage[final]{acl}

\usepackage{microtype}

\usepackage[english,bidi=default]{babel}
\babelfont{rm}{TeXGyreTermesX}
\babelprovide[import]{hindi}
\babelfont[*devanagari]{rm}{Lohit Devanagari}
\babelprovide[import]{arabic}
\babelfont[*arabic]{rm}{Noto Sans Arabic}

\usepackage{fontspec}
\newfontfamily\ipafont[
  Scale=MatchLowercase,
  UprightFont    = CharisSIL-Regular.ttf,
  ItalicFont     = CharisSIL-Italic.ttf,
  BoldFont       = CharisSIL-Bold.ttf,
  BoldItalicFont = CharisSIL-BoldItalic.ttf
]{CharisSIL-Regular.ttf}
\DeclareTextFontCommand{\textipa}{\ipafont}

\usepackage{booktabs}
\usepackage{siunitx}
\usepackage{graphicx}
\usepackage{subcaption}
\usepackage{amsmath}
\usepackage{amsfonts}
\usepackage{xurl}
\usepackage{placeins}
\usepackage{fontawesome5}

\usepackage{tikz}
\usetikzlibrary{arrows.meta, positioning}
\usepackage[dvipsnames]{xcolor}
\definecolor{text}{HTML}{538AE5}
\definecolor{ipa}{HTML}{F0434F}
\definecolor{textsub}{HTML}{A9C4F0}
\definecolor{ipasub}{HTML}{F9A1A8}
\definecolor{40k}{HTML}{9D8977}
\definecolor{80k}{HTML}{886E58}
\definecolor{100k}{HTML}{735238}
\definecolor{200k}{HTML}{5E3719}

\usepackage{pifont}
\newcommand{\cmark}{\textcolor{ForestGreen}{\ding{51}}}

\usepackage{calc}
\newcommand{\erceulogo}{\raisebox{-0.1\height}{\includegraphics[height=6ex]{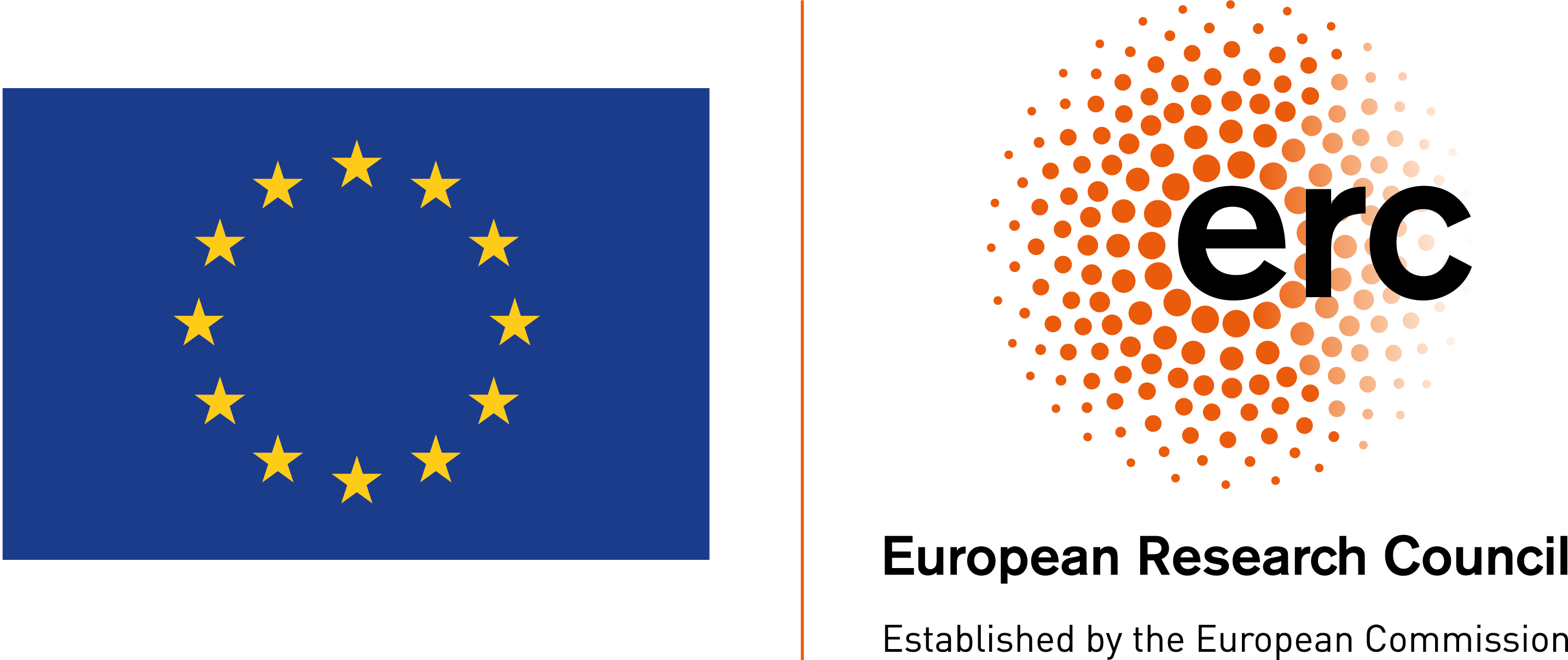}}}

\usepackage{fnpct}
\usepackage{multirow}

\title{Phonemes to the Rescue:\\
Multilingual Tokenization Based on International Phonetic Alphabet}

\author{Milan Mileti\'{c} \\
  University of Amsterdam \\
  \texttt{m.miletic@uva.nl} \\\And
  Julie Kallini \\
  Stanford University \\
  \texttt{kallini@stanford.edu} \\\And
  Ekaterina Shutova \\
  University of Amsterdam \\
  \texttt{e.shutova@uva.nl}
  }

\begin{document}

\maketitle
\begin{abstract}
Multilingual language models often exhibit performance disparities across languages that can arise as early as the tokenization stage. Widely-used subword tokenization approaches favor high-resource languages, and tokenizer-free methods still yield longer sequences for scripts with a higher bytes-per-character ratio. To address these shortcomings, we propose to use the International Phonetic Alphabet (IPA) as a language-agnostic input representation for multilingual tokenizers. IPA provides a compact symbol inventory, greater cross-lingual character overlap, and a more balanced byte-per-character distribution across languages. We train matched pairs of text vs. IPA subword tokenizers across 24 languages and 14 scripts and demonstrate that IPA tokenizers consistently improve tokenization quality, especially for non-Latin scripts, and generalize more effectively to unseen languages and scripts.
\end{abstract}

\begin{center}
    \small \faGithub\ \href{https://github.com/Mikki99/ipa-tokenization}{github.com/Mikki99/ipa-tokenization}
\end{center}

\section{Introduction}

Despite their growing global impact, multilingual language models (MLMs) continue to exhibit performance disparities across languages. Prior work shows that these differences originate as early as the tokenization stage, with direct consequences for downstream model performance \citep{petrov2023language, lotz-etal-2025-beyond, arnett2025language, rust2020good}.
\begin{figure}[t]
  \centering
  \includegraphics[width=\columnwidth]{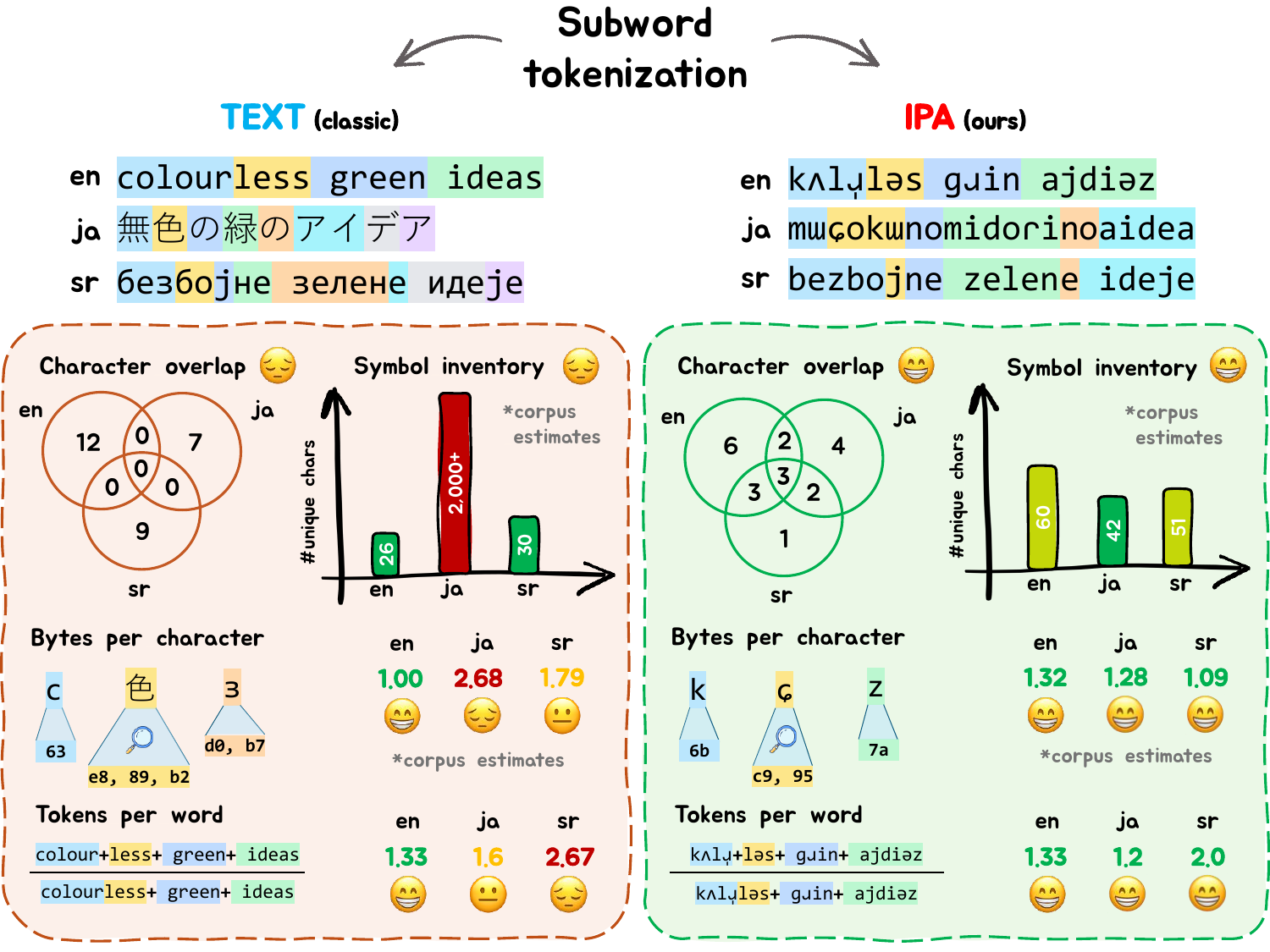}
  \caption{Benefits of IPA for multilingual tokenization. We show how our trained \textcolor{text}{\textbf{Text}} and \textcolor{ipa}{\textbf{IPA}} tokenizers segment the same input sequence in English (en), Japanese (ja), and Serbian (sr). Switching to IPA maps all languages into a shared phonetic alphabet, which (i) increases cross-lingual character overlap, (ii) reduces the symbol inventory needed to represent the corpus, and (iii) mitigates byte-per-character differences. Together, this results in better overall tokenization quality, as shown for instance by reduced tokens per word (what we later refer to as word fertility).}
  \label{fig:figure1}
\end{figure}
The main challenge of multilingual tokenization is how to represent a large number of diverse languages in a common fixed-size vocabulary. Currently predominant subword tokenization algorithms (BPE, \citealt{sennrich-etal-2016-bpe}; UnigramLM, \citealt{kudo2018unigram}) learn this vocabulary by optimizing objectives defined over corpus statistics. By design, this process favors languages that are more prevalent in the training set. As a result, underrepresented languages, especially those with rich morphologies or those using non-Latin scripts, often require more tokens for encoding the same semantic content \citep{ahia2023all}. This leads to increased computational cost, slower inference times, and diminished fluency and output quality in those languages \citep{qin2025survey}. A prominent alternative to subword tokenization in recent years is \emph{tokenization-free} byte-level modeling, which represents each input byte as a token. While this approach avoids many of the issues associated with traditional tokenizers, the shift to byte tokens \textit{alone} does not eliminate cross-lingual disparities: characters in some scripts require up to 4$\times$ as many bytes to encode than those in other scripts, leading to longer input sequences---and thus, higher computational costs---for those languages \citep{mielke2021between,limisiewicz-etal-2024-myte}. This indicates that tokenization quality across languages depends not only on the tokenization method (or lack thereof), but also on the input encoding on which it operates.

Guided by this observation, in this work we consider an alternative input representation that is offered by the International Phonetic Alphabet (IPA, \citealt{ipa1999}) instead of using standard orthography. IPA maps languages into a shared (phonetic) alphabet, increasing character-level overlap across languages that otherwise use disjoint scripts. Because the symbol inventory is small ($\approx 200$ characters), a fixed-size subword vocabulary can be used more efficiently, with less of its capacity wasted on script-specific variants of the same words or morphemes. Finally, IPA is encoded mostly with 1–2 byte UTF-8 characters, which mitigates script-driven differences in bytes-per-character and therefore reduces disparities in per-language encoding cost. An overview of IPA's benefits is illustrated in Fig.~\ref{fig:figure1} and we provide empirical results to support these claims in Appendix~\ref{apdx:ipa_benefits}. These benefits inspired IPA's usage in NLP at various stages of the modeling pipeline, from pre-training tasks \citep{gale2023mixed} to prompting \citep{nguyen2024prompting}, or even integrating it into multimodal pipelines \citep{matsuhira2023ipa}. However, to the best of our knowledge, we are the first to systematically evaluate the effects of IPA on multilingual tokenization quality across a diverse set of configurations and languages.   

Concretely, we train subword tokenizers on multilingual data spanning 24 languages and 14 distinct scripts in two input formats---standard orthography and IPA---while varying the tokenization algorithm, vocabulary size, and data sampling strategy. We evaluate each tokenizer on a suite of intrinsic tokenization quality metrics measuring compression, token frequency distribution properties, vocabulary usage, and cross-lingual equity. We further train GPT-2 models on a set of selected tokenizers and evaluate them on two widely used multilingual downstream tasks: XNLI and PAWS-X. We find that IPA tokenizers consistently outperform standard text tokenizers on intrinsic metrics, under their respective best configurations, without sacrificing downstream task performance. Overall, our approach preserves MLM capabilities, while yielding more equitable treatment across languages.

\section{Background and Related Work}\label{sec:rel_work}

\begin{table*}[t]
\centering
\scriptsize
\setlength{\tabcolsep}{2.2pt}
\renewcommand{\arraystretch}{1.15}
\resizebox{\textwidth}{!}{%
\begin{tabular}{lccccccccc ccc cc c c c c c c c c c c c}
\toprule
& \multicolumn{9}{c}{\textsc{Latn}} & \multicolumn{3}{c}{\textsc{Arab}} & \multicolumn{2}{c}{\textsc{Cyrl}} & \multicolumn{1}{c}{\textsc{Deva}} & \multicolumn{1}{c}{\textsc{Jpan}} & \multicolumn{1}{c}{\textsc{Hang}} & \multicolumn{1}{c}{\textsc{Hani}} & \multicolumn{1}{c}{\textsc{Thai}} & \multicolumn{1}{c}{\textsc{Laoo}} & \multicolumn{1}{c}{\textsc{Mymr}} & \multicolumn{1}{c}{\textsc{Grek}} & \multicolumn{1}{c}{\textsc{Hebr}} & \multicolumn{1}{c}{\textsc{Ethi}} \\
\cmidrule(lr){2-10}\cmidrule(lr){11-13}\cmidrule(lr){14-15}\cmidrule(lr){16-16}\cmidrule(lr){17-17}\cmidrule(lr){18-18}\cmidrule(lr){19-19}\cmidrule(lr){20-20}\cmidrule(lr){21-21}\cmidrule(lr){22-22}\cmidrule(lr){23-23}\cmidrule(lr){24-24}\cmidrule(lr){25-25}\cmidrule(lr){26-26}
\textbf{Stage}
& \textsc{de} & \textsc{en} & \textsc{es} & \textsc{fi} & \textsc{fr} & \textsc{it} & \textsc{pl} & \textsc{sw} & \textsc{tr}
& \textsc{ar} & \textsc{fa} & \textsc{ur}
& \textsc{ru} & \textsc{sr}
& \textsc{hi}
& \textsc{ja}
& \textsc{ko}
& \textsc{zh}
& \textsc{th}
& \textsc{lo}
& \textsc{my}
& \textsc{el}
& \textsc{he}
& \textsc{am} \\
\midrule
Tokenizer training
& \cmark & \cmark & - & \cmark & \cmark & \cmark & - & \cmark & \cmark
& \cmark & \cmark & \cmark
& \cmark & \cmark
& \cmark
& \cmark
& \cmark
& \cmark
& \cmark
& \cmark
& -
& -
& -
& - \\
GPT-2 pre-training
& \cmark & \cmark & - & \cmark & \cmark & \cmark & - & \cmark & \cmark
& \cmark & \cmark & \cmark
& \cmark & \cmark
& \cmark
& \cmark
& \cmark
& \cmark
& \cmark
& \cmark
& -
& -
& -
& - \\
Intrinsic eval
& \cmark & \cmark & \cmark & \cmark & \cmark & \cmark & \cmark & \cmark & \cmark
& \cmark & \cmark & \cmark
& \cmark & \cmark
& \cmark
& \cmark
& \cmark
& \cmark
& \cmark
& \cmark
& \cmark
& \cmark
& \cmark
& \cmark \\
XNLI
& \cmark & \cmark & \cmark & - & \cmark & - & - & \cmark & \cmark
& \cmark & - & \cmark
& \cmark & -
& \cmark
& -
& -
& \cmark
& \cmark
& -
& -
& \cmark
& -
& - \\
PAWS-X
& \cmark & \cmark & \cmark & - & \cmark & - & - & - & -
& - & - & -
& - & -
& -
& \cmark
& \cmark
& \cmark
& -
& -
& -
& -
& -
& - \\
\bottomrule
\end{tabular}%
}
\caption{Language coverage at different stages of our experiments. Columns show languages (ISO 639-1 codes) grouped by script (ISO 15924 codes). We provide the language/script-to-ISO mapping in Appendix~\ref{apdx:iso_mappings}. \cmark\ indicates that a language is included in the corresponding stage. For tokenizer and GPT-2 training we use a sample from the CulturaX dataset \citep{nguyen2023culturax}. For intrinsic evaluation we use WikiPron \citep{lee2020wikipron} and FLORES+ \citep{nllb2024scaling}. We perform downstream evaluation on XNLI \citep{conneau-etal-2018-xnli} and PAWS-X \citep{yang-etal-2019-paws}.}
\label{tab:lang_coverage_matrix}
\end{table*}

\paragraph{Disparities in tokenization quality.} Prior work has shown that standard multilingual subword tokenizers systematically treat languages inequitably \citep[][inter alia]{petrov2023language, ahia2023all, ali2024tokenizer}. Resulting frequency-based vocabularies allocate more capacity to high-resource languages, so that lower-resource languages are split into many more tokens per word. This is especially amplified in case of morphologically more complex languages, as well as those using non-Latin scripts. \citet{petrov2023language} quantify this disparity, showing a difference in the amount of tokens needed to encode the same semantic content of up to $15\times$ between languages. This over-segmentation has direct consequences, as the impacted languages will require much higher training and inference costs\footnote{\citet{lundin2025token} refer to this effect as \textit{token tax}, showing that doubling in tokens can increase training cost by a factor of four.}. 

These findings have inspired several recent works that aim to tackle these tokenization challenges. \citet{petrov2023language} suggest training monolingual tokenizers for each language and then merging them based on tokenization parity. Inheriting the idea of tokenization parity, \citet{foroutan2025parity} introduce Parity-Aware BPE, optimizing the training objective in classic BPE such that merges are guided by per-language compression levels, instead of simple frequency. \citet{feher2025retrofitting} propose a dynamic tokenization approach, which merges frequently co-occurring subwords on the fly to reduce the token-count inflation. The main alternative to subword tokenization in recent years have been different flavors of byte-level approaches, which eliminate learned segmentation by treating each byte in the input as a separate token \citep[][inter alia]{xue2022byt5, kallini2024mrt5, yu2023megabyte}. However, cross-lingual disparities still remain even when working with byte tokens, as languages differ in the number of bytes-per-character, as well as the number of symbols needed to encode the same semantic content \citep{arnett2024bit}. Recent work has sought to address this issue with more adaptive forms of byte-level tokenization. \citet{ahia2024magnet} introduce \textsc{MAGNET}, which integrates adaptive gradient-based tokenization into the model through script-specific predictors that place token boundaries so as to equalize segmentation granularity across scripts. \citet{owodunni2025flexitokens} propose \textsc{FlexiTokens}, introducing a more flexible training objective that allows the compression rate to adapt across languages and scripts, as well as across domains within the same language.

While our method could in principle be applied to byte-level approaches (see Appendix~\ref{apdx:ipa_byte_level}), we focus on currently predominant subword tokenization in this work. We provide a summary of different tokenization methods in Appendix~\ref{apdx:tok}.

\paragraph{IPA in Natural Language Processing.} IPA has seen a growing range of applications in NLP in recent years, particularly in settings that require phonological awareness. \citet{leong2022phone} and \citet{sohn2025cross} show benefits of IPA representations for improved performance in named entity recognition (NER) tasks. \citet{gale2023mixed} introduce the BORT model, an extension of BART \citep{lewis2020bart} with added self-supervised pre-training pronunciation tasks by leveraging IPA representations. \citet{nguyen2024prompting} propose phonemic prompting as a way to enhance the multilingual capabilities of LLMs. \citet{matsuhira2023ipa} developed IPA-CLIP, by incorporating IPA representations into a multimodal setting, extending the CLIP model. Taken together, these studies highlight the versatility and effectiveness of IPA in enhancing cross-lingual generalization, representation learning, and multilingual performance. Most closely related to our work are \citet{goriely2024babble}, who also pre-train GPT-2 models on IPA input, testing both character-level and BPE tokenization. However, their study is limited to a monolingual setting (English) and does not analyze the impact of IPA on tokenization quality.

\section{Data and Languages}
\label{sec:data_langs}

\subsection{Languages}
\label{sec:languages}

Table~\ref{tab:lang_coverage_matrix} summarizes language coverage at each stage of our experiments. For tokenizer training (Section~\ref{sec:tokenizer_training}) and GPT-2 pre-training (Section~\ref{sec:gpt2}) we consider 18 languages spanning 10 scripts. We evaluate the intrinsic tokenization quality (Section~\ref{sec:intrinsic}) on the same language set together with six additional languages (\textsc{am}, \textsc{el}, \textsc{es}, \textsc{he}, \textsc{my}, \textsc{pl}) that introduce four new scripts (\textsc{Grek}, \textsc{Hebr}, \textsc{Ethi}, \textsc{Mymr}) and allow us to test zero-shot generalization to languages/scripts unseen during tokenizer training. In total, this results in 24 languages and 14 scripts considered in this work.

Our language selection aimed to cover a diverse sample across orthographic, typological, and resource profiles. For instance, it includes high-resource Indo-European languages (e.g., \textsc{en}, \textsc{de}, \textsc{fr}, \textsc{it}, \textsc{ru}), as well as comparatively lower-resource languages (e.g., \textsc{sw}, \textsc{lo}, \textsc{my}, \textsc{am}), which are often underrepresented in multilingual pre-training pipelines. We also consider morphologically rich languages (e.g., \textsc{fi}, \textsc{tr}) which often exhibit worse tokenization quality \citep{raj2024every} and those with more complex orthographies (e.g., \textsc{zh}, \textsc{ja}, \textsc{ko}, \textsc{th}, \textsc{lo}) which are typically more expensive to encode due to their Unicode ranges. We also include a language written in multiple scripts, \textsc{ja}, spanning Katakana, Hiragana, and Kanji (collectively labeled as \textsc{Jpan})\footnote{Although, in our experiments we only consider Katakana and Hiragana, due to limitations in Epitran's Kanji support}\footnote{Serbian is also a multi-script language (\textsc{Latn} and \textsc{Cyrl}), but our sample only contained Cyrillic text.}.

\subsection{Datasets}
\label{sec:datasets}

\paragraph{CulturaX.} For tokenizer training (Section~\ref{sec:tokenizer_training}) and GPT-2 pre-training (Section~\ref{sec:gpt2}), we use CulturaX \citep{nguyen2023culturax}, a large-scale multilingual web corpus derived from filtered mC4 \citep{xue2021mt5massivelymultilingualpretrained} and OSCAR \citep{suarez2019oscar1,suarez2020oscar2}. We apply additional post-processing to remove residual non-linguistic artifacts and reduce cross-language contamination on our sample of languages (Appendix~\ref{app:culturax_cleaning}). We sample a total of 1GB of data distributed across our 18 languages using four strategies:
(i) \textcolor{JungleGreen}{\texttt{byte-uniform}}, allocating an equal number of bytes to each language;
(ii) \textcolor{LimeGreen}{\texttt{semantic-uniform}}, allocating equal semantic content by compensating for language-specific byte premiums \citep{arnett2024bit};
(iii) \textcolor{BrickRed}{\texttt{data-proportional}}, sampling in proportion to each language’s share in CulturaX; and
(iv) \textcolor{RedOrange}{\texttt{data-smoothed}}, applying temperature sampling ($\alpha{=}0.3$) to upweight lower-resource languages (as done in mT5, \citet{xue2021mt5massivelymultilingualpretrained}; mBERT, \citet{devlin2018mbert}; XLM-R, \cite{conneau2020unsupervised}). The first two strategies provide a \emph{balanced} sample of languages (under different notions of balance), while the latter two result in \emph{unbalanced}, but arguably more realistic distributions in practice. More details and distribution visualizations are in Appendix~\ref{apdx:sampling_strategies}.

\paragraph{WikiPron and FLORES+.} For evaluating intrinsic tokenization quality we use two datasets: (i) WikiPron \citep{lee2020wikipron}, a multilingual collection of word lists paired with IPA transcriptions extracted from Wiktionary\footnote{\url{https://www.wiktionary.org/}} and (ii) FLORES+ \citep{nllb2024scaling}, a dataset of human-translated parallel sentences across more than 200 languages. We use WikiPron for word-level metrics (word fertility and proportion of continued words; see Section~\ref{sec:intrinsic}) and FLORES+ for all the remaining metrics, computed at the sentence level. Details can be found in Appendix~\ref{apdx:data_and_langs}.

\paragraph{XNLI and PAWS-X.} We fine-tune and evaluate selected models on XNLI \citep{conneau-etal-2018-xnli} and PAWS-X \citep{yang-etal-2019-paws}, two standard multilingual natural language understanding (NLU) benchmarks. These two benchmarks cover 13 and 7 languages from our selection, respectively, as shown in Table~\ref{tab:lang_coverage_matrix}.

\section{Methodology}
\label{sec:method}

\subsection{Overview}
\label{sec:method_overview}
Our experiments compare subword tokenization trained on two input representations: standard orthographic text (\textcolor{text}{\textbf{Text}}) and its phonemic transcription in the International Phonetic Alphabet (\textcolor{ipa}{\textbf{IPA}}). Starting from the same multilingual corpora, we (i) build an IPA version of each dataset with a multilingual grapheme-to-phoneme (G2P) pipeline, (ii) train matched pairs of \textcolor{text}{\textbf{Text}} vs. \textcolor{ipa}{\textbf{IPA}} tokenizers on different configurations, (iii) evaluate intrinsic tokenization quality across languages, and (iv) pre-train GPT-2 models with selected tokenizers and evaluate cross-lingual transfer by fine-tuning on English and testing zero-shot on other languages.

\subsection{Grapheme-to-phoneme conversion}
\label{sec:g2p}
To train and evaluate IPA tokenizers and IPA-based language models, we require IPA text at scale. While some resources provide curated IPA transcriptions (e.g., WikiPron \citep{lee2020wikipron} or Universal Dependencies (UD; \citeauthor{nivre2020ud}, \citeyear{nivre2020ud})), these are often limited in size and language coverage. In this work, we utilize Epitran \citep{mortensen2018epitran}, an open-source rule-based grapheme-to-phoneme (G2P) toolkit designed specifically for multilingual applications. Epitran supports more than 60 languages across a variety of scripts and is being actively extended. It takes standard orthographic text as input, together with the source language and script code (e.g. \textcolor{text}{\texttt{tokenization}}, \textcolor{violet}{\texttt{eng-Latn}}), and returns its IPA equivalent (e.g. \textcolor{ipa}{\textipa{towkənəzejʃən}}). We applied minor language-specific fixes and extensions, as well as efficiency improvements to Epitran's conversion pipeline, to adapt it best to our needs (see Appendix~\ref{apdx:epitran} for a detailed discussion). For languages that are not supported by Epitran (Greek and Hebrew), we use Phonemizer \citep{Bernard2021} instead. We discuss G2P conversion quality in Appendix~\ref{apdx:g2p_quality}

\subsection{Tokenizer training}
\label{sec:tokenizer_training}
To isolate how the input representation affects subword tokenization we vary a set of configurations outlined below. For each configuration we train a paired set of tokenizers: one on \textcolor{text}{\textbf{Text}} and one on \textcolor{ipa}{\textbf{IPA}}. We train all tokenizers with SentencePiece \citep{kudo2018sentencepiece} to allow comparisons across subword algorithms under consistent training conditions and because it avoids whitespace-based pre-tokenization, which is important for languages in our study without explicit word boundary markers. We detail hyper-parameter settings and representation-specific adjustments (such as Unicode normalization and character coverage) in Appendix~\ref{apdx:tok_HPs}.

\paragraph{Experimental grid.} We vary three factors in our experiments: (1) subword tokenization algorithm (\textcolor{orange}{BPE}, \textcolor{violet}{UnigramLM}), (2) vocabulary size (\textcolor{40k}{40k}, \textcolor{80k}{80k}, \textcolor{100k}{100k}, \textcolor{200k}{200k}), and (3) data sampling strategy (\textcolor{JungleGreen}{\texttt{byte-uniform}}, \textcolor{LimeGreen}{\texttt{semantic-uniform}}, \textcolor{BrickRed}{\texttt{data-proportional}}, \textcolor{RedOrange}{\texttt{data-smoothed}}; defined in Section~\ref{sec:tokenizer_training}). This yields $2 \times 4 \times 4 = 32$ configurations, and thus 32 paired \textcolor{text}{\textbf{Text}}/\textcolor{ipa}{\textbf{IPA}} tokenizers (64 tokenizers in total).

\subsection{Intrinsic evaluation metrics}
\label{sec:intrinsic}
We assess tokenization quality using ten metrics grouped into four categories---compression, token frequency distribution shape, vocabulary usage, and cross-lingual equity---and refer to Appendix~\ref{apdx:intrinsic_metrics} for formal definitions.

\subsubsection{Compression}
We report four compression metrics: (i) \textbf{Word Fertility (WF)}, the average number of subword tokens per unique word type, where lower values indicate that words are represented with fewer pieces; (ii) \textbf{Proportion of Continued Words (PCW)}, the fraction of unique words split into more than one token, where lower values indicate fewer splits; (iii) \textbf{Average Token Length (ATL)}, the mean token length measured in input characters of the representation (\textcolor{text}{\textbf{Text}} or \textcolor{ipa}{\textbf{IPA}}), where higher values indicate tokens spanning longer segments; and (iv) \textbf{Compression Rate (CR)}, the average number of input characters per token at the sentence level, where higher values indicate stronger compression.

\subsubsection{Token frequency distribution shape}
We characterize the token frequency profile with two metrics: (i) \textbf{R\'enyi Entropy (RE)}, computed from token frequencies on evaluation data and reported for $\alpha{=}1$ (Shannon), $\alpha{=}2$ (collision), and $\alpha{=}\infty$ (min-entropy), where higher values indicate more uniform token usage across the vocabulary; and (ii) \textbf{Zipf Deviation (ZipfD)}, which measures how closely the empirical rank--frequency curve matches an ideal Zipfian distribution, where lower values indicate closer adherence to Zipf's law \citep{lotz-etal-2025-beyond}.

\subsubsection{Vocabulary usage}
We quantify vocabulary usage with two metrics: (i) \textbf{Vocabulary Utilization (VU)}, the share of the learned vocabulary that appears at least once in the evaluation data, where higher values indicate less unused capacity; and (ii) \textbf{Type--Token Ratio (TTR)}, the number of distinct token types divided by the total number of produced tokens, where higher values indicate more diverse token usage relative to sequence length.

\begin{figure*}[t]
  \centering
  \includegraphics[width=2.1\columnwidth]{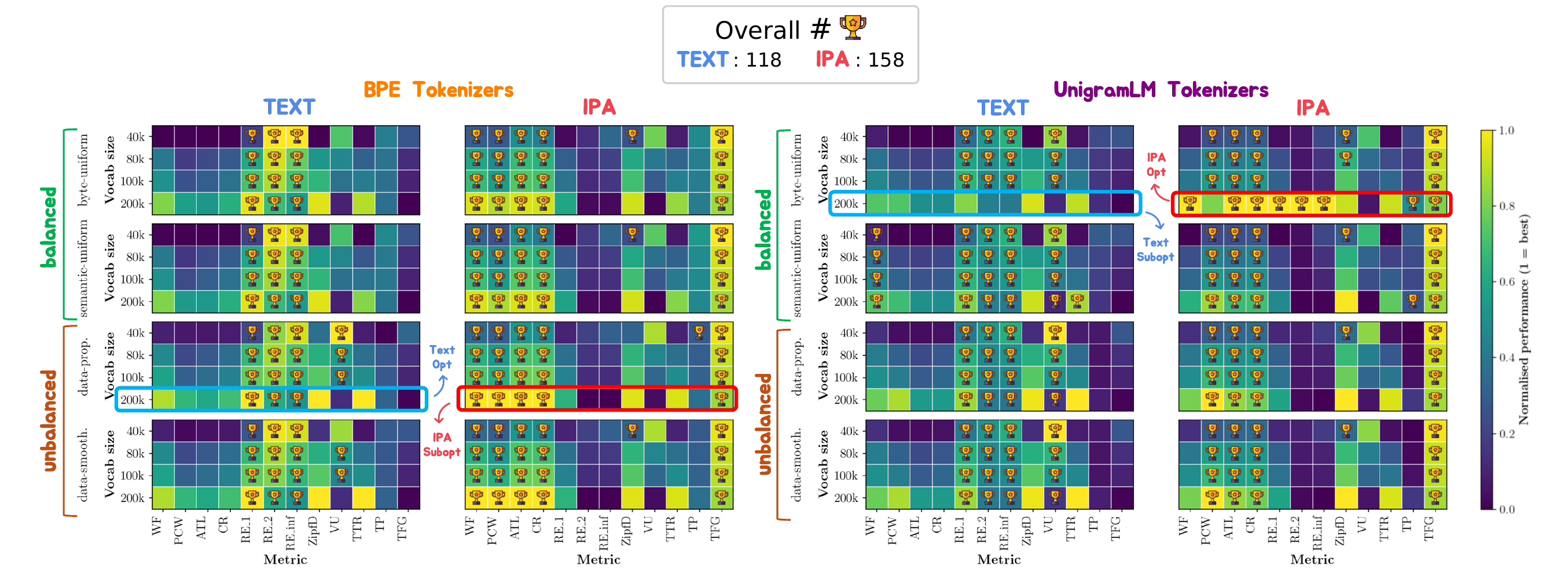}
  \caption{Intrinsic performance across all tokenizer configurations for \textcolor{text}{\textbf{Text}} and \textcolor{ipa}{\textbf{IPA}} representations. Each heatmap cell reports a metric value macro-averaged across languages and min--max normalized to $[0,1]$ (higher is better) for a configuration defined by vocabulary size, subword algorithm, and data-sampling strategy. Cells marked with a trophy indicate a ${>}0.1$ advantage over the paired configuration in the other representation (\textcolor{text}{\textbf{Text}} vs.\ \textcolor{ipa}{\textbf{IPA}}). Outlined cells denote the selected \emph{Opt/Subopt} tokenizers for GPT-2 training. \textbf{Takeaway:} Across configurations, \textcolor{ipa}{\textbf{IPA}} is consistently stronger on compression (WF, PCW, ATL, CR) and cross-lingual equity (TFG), while \textcolor{text}{\textbf{Text}} yields higher token-frequency uniformity as measured by R\'enyi entropies ($\alpha\in\{1,2,\infty\}$). Overall, \textcolor{ipa}{\textbf{IPA}} wins more often (158 trophies) than \textcolor{text}{\textbf{Text}} (118 trophies).}
  \label{fig:heatmap}
\end{figure*}

\subsubsection{Cross-lingual equity}
We assess cross-lingual equity with two metrics: (i) \textbf{Tokenization Parity (TP)}, the average ratio of token counts between English and a target language computed over parallel sentence pairs, where values closer to $1$ indicate similar segmentation length \citep{petrov2023language}; and (ii) \textbf{Tokenization Fairness Gini (TFG)}, the Gini coefficient over language-level tokenization cost, where cost is defined as token count normalized by input bytes, and lower values indicate more equal tokenization cost across languages \citep{meister_tokenizer_analysis_2025}.

\begin{figure*}[t]
  \centering
  \includegraphics[width=2\columnwidth]{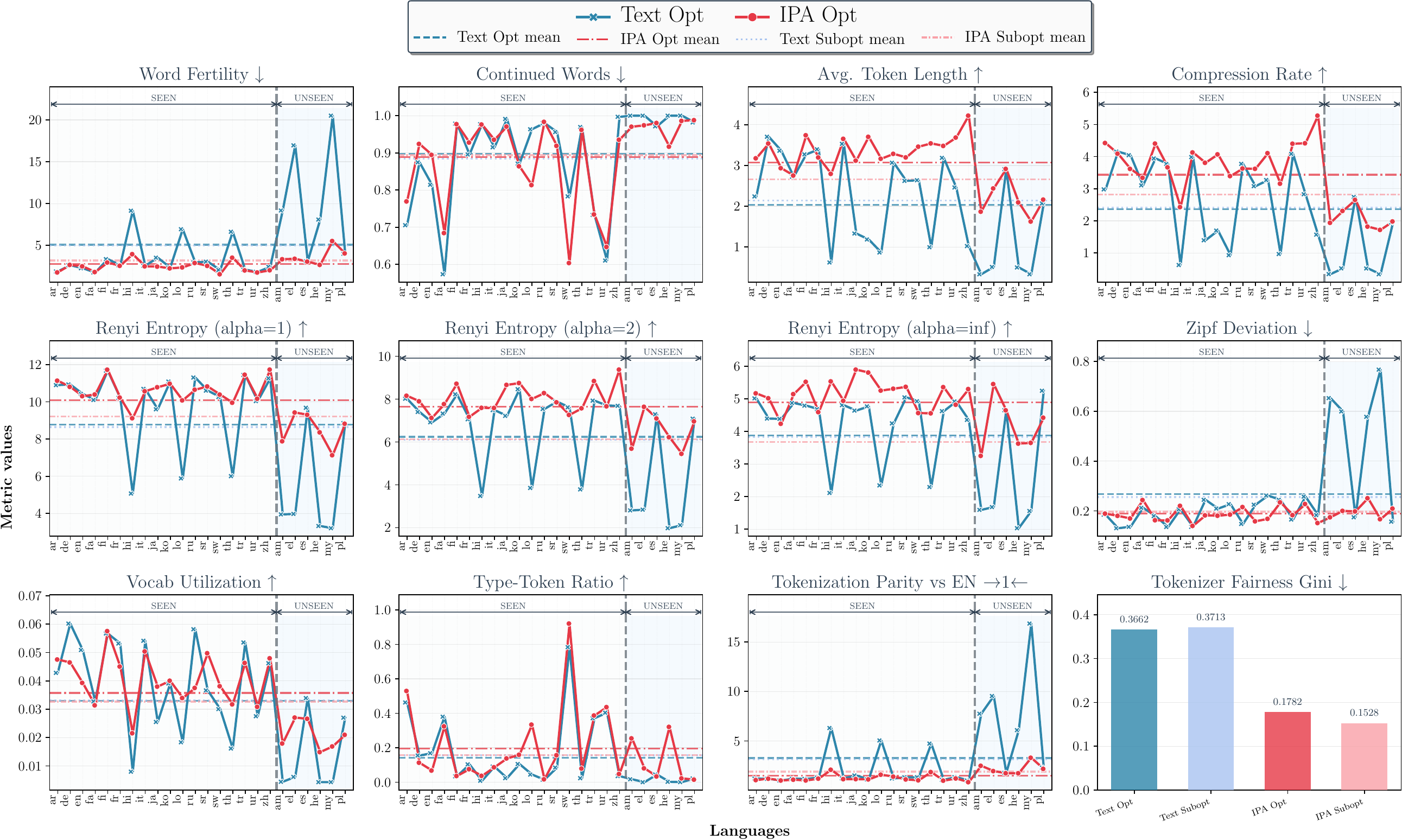}
  \caption{Per-language intrinsic tokenization metrics: \textcolor{text}{\textbf{Text}} vs.\ \textcolor{ipa}{\textbf{IPA}}. Solid lines show \emph{Text Opt} and \emph{IPA Opt} per language; dashed lines are means. \emph{Subopt} variants shown as mean-only (per-language in Appendix~\ref{apdx:subopt_per_lang}). Arrows: $\uparrow$ = higher is better, $\downarrow$ = lower is better, $\rightarrow1\leftarrow$ = closer-to-1 is better; TFG is a single global tokenizer metric (not per-language), hence shown as bars. \textsc{Seen}/\textsc{Unseen} mark languages included in vs.\ absent from tokenizer training data. \textbf{Takeaway:} \emph{IPA (Opt)} shows consistently better tokenization quality than \emph{Text (Opt and Subopt)} across metrics, with differences particularly pronounced for unseen languages.}
  \label{fig:intrinsic_per_lang}
\end{figure*}

\subsection{GPT-2 pre-training and fine-tuning}
\label{sec:gpt2}

\paragraph{Tokenizer selection.} Since training a language model for all 32 tokenizer settings for both Text and IPA is computationally expensive, we select a small set of tokenizers for GPT-2 pre-training experiments using our intrinsic evaluation. We rank the 32 Text and 32 IPA tokenizers by macro-averaging each intrinsic metric over languages and then aggregating the resulting metric-wise ranks into a single mean-rank score. We select the highest ranked Text and IPA tokenizers, which we refer to as \textcolor{text}{\emph{\textbf{Text Opt}}} and \textcolor{ipa}{\emph{\textbf{IPA Opt}}}, respectively. \textcolor{text}{\emph{\textbf{Text Opt}}} tokenizer is found under the following setting [\textcolor{orange}{BPE}, \textcolor{200k}{200k}, \textcolor{BrickRed}{\texttt{data-proportional}}], while for \textcolor{ipa}{\emph{\textbf{IPA Opt}}} it is [\textcolor{violet}{UnigramLM}, \textcolor{200k}{200k}, \textcolor{JungleGreen}{\texttt{byte-uniform}}]. Because these optima arise under different settings, the comparison is no longer driven only by the input representation. Therefore, we also select two control tokenizers: Text tokenizer with IPA Opt settings (\textcolor{textsub}{\emph{\textbf{Text Subopt}}}) and IPA tokenizer with Text Opt settings (\textcolor{ipasub}{\emph{\textbf{IPA Subopt}}}). We, thus, pre-train four GPT-2 models in total. The ranking procedure and the full ranking results are reported in Appendix~\ref{apdx:tok_selection}.

\paragraph{Pre-training.} We train GPT-2 Small models \citep{radford2018improving, radford2019languagemodels} from scratch on the \textcolor{RedOrange}{\texttt{data-smoothed}} sample of CulturaX for our 18 languages using the four selected tokenizers. Given that all of our tokenizers use a $200$k vocabulary, we end up with a total of 240M parameter models (rather than 125M in the original GPT-2 Small, which uses a $50$k vocabulary). All hyperparameters are fixed across models and their values are reported in Appendix~\ref{apdx:gpt2_HPs}.

\paragraph{Fine-tuning and evaluation.} For both XNLI and PAWS-X, we consider three fine-tuning regimes: (1) monolingual, (2) multilingual, and (3) English-only. In the \textit{monolingual} regime, a pre-trained model is fine-tuned and evaluated independently for each language. In the \textit{multilingual} regime, the model is fine-tuned jointly on data from all languages and evaluated separately on each language. In the \textit{English-only} regime, the model is fine-tuned exclusively on the English training data and evaluated both on a held-out English split and, in a zero-shot setting, on all remaining languages to measure cross-lingual transfer abilities.

\section{Results and Analysis}
\begin{figure}[t]
  \centering
  \begin{subfigure}[t]{0.49\textwidth}
    \centering
    \includegraphics[width=\linewidth]{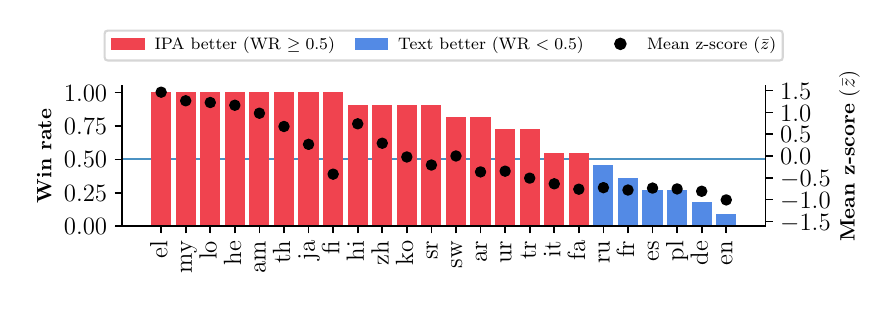}
    \caption{Per-language}
    \label{fig:per_lang_winrate}
  \end{subfigure}\hspace{0.02\textwidth}
  \begin{subfigure}[t]{0.49\textwidth}
    \centering
    \includegraphics[width=\linewidth]{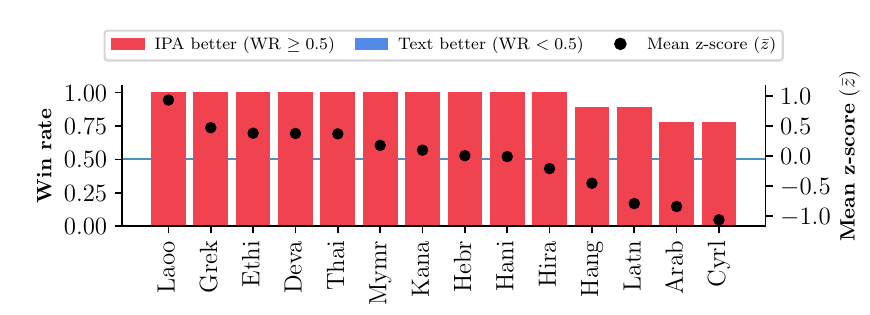}
    \caption{Per-script}
    \label{fig:per_script_winrate}
  \end{subfigure}
  \caption{Ranking \textcolor{ipa}{\textbf{IPA}} impact on tokenization quality (a) per-language and (b) per-script. Bars show win rate (WR), the fraction of intrinsic metrics on which \emph{IPA Opt} outperforms \emph{Text Opt}: $\mathrm{WR}\!\in\![0,1]$, with $\mathrm{WR}{=}1/0$ indicating that \textcolor{ipa}{\textbf{IPA}}/\textcolor{text}{\textbf{Text}} wins on all metrics, respectively. \textcolor{ipa}{\textbf{Red}} bars mark languages/scripts where \textcolor{ipa}{\textbf{IPA}} wins on a majority of metrics ($\mathrm{WR}{>}0.5$), while \textcolor{text}{\textbf{blue}} bars mark those where \textcolor{text}{\textbf{Text}} wins ($\mathrm{WR}{<}0.5$). Black dots show the \textbf{mean z-score} $\bar{z}$ (average standardized IPA--Text delta; $\bar{z}{>}0$ favors IPA). \textbf{Takeaway:} The largest improvements concentrate on unseen scripts (\textsc{Grek}, \textsc{Ethi}, \textsc{Mymr}, \textsc{Hebr}) and on languages with more complex orthographies (e.g., \textsc{lo, th, ja}), while only a small set of mostly high-resource Latin languages (\textsc{fr, es, pl, de, en}) and one Cyrillic language (\textsc{ru}) show lower overall quality.}
  \label{fig:winrate_with_z}
\end{figure}

\begin{figure}[t]
  \centering
  \includegraphics[width=\columnwidth]{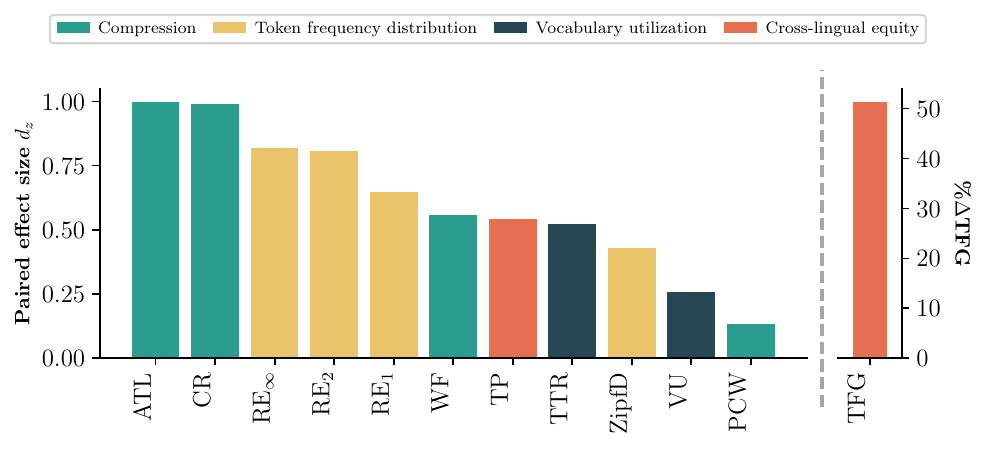}
  \caption{Ranking \textcolor{ipa}{\textbf{IPA}} impact on tokenization quality per metric. Bars are grouped and color-coded by metric family (see Sec.~\ref{sec:intrinsic}). Left panel reports paired effect size $d_z$ across languages for 11 per-language metrics ($d_z>0$ favors \textcolor{ipa}{\textbf{IPA}}, $d_z<0$ favors \textcolor{text}{\textbf{Text}}). TFG is shown separately (right panel) because it is a single global tokenizer statistic rather than a per-language metric; we report its relative change ($\%\Delta\mathrm{TFG}_{\textcolor{text}{\mathrm{Text\,Opt}}-\textcolor{ipa}{\mathrm{IPA\,Opt}}}$; $>0$ favors \textcolor{ipa}{\textbf{IPA}}). \textbf{Takeaway:} IPA improves average intrinsic quality across all metrics, with the largest gains in compression (ATL, CR) and cross-lingual equity (TFG), suggesting better tokenization efficiency and more uniform cross-lingual behavior.}
  \label{fig:metric_impact}
\end{figure}

\begin{figure*}[htbp]
    \centering
    
    \begin{subfigure}{\textwidth}
        \centering
        \includegraphics[width=\linewidth]{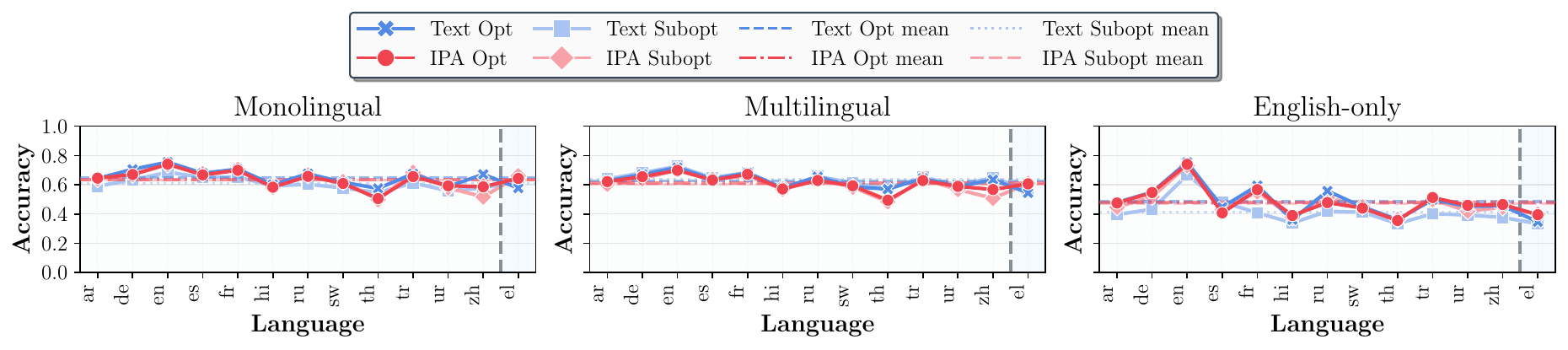}
        \caption{XNLI.}
        \label{fig:downstream_xnli}
    \end{subfigure}
    
    \begin{subfigure}{\textwidth}
        \centering
        \includegraphics[width=\linewidth]{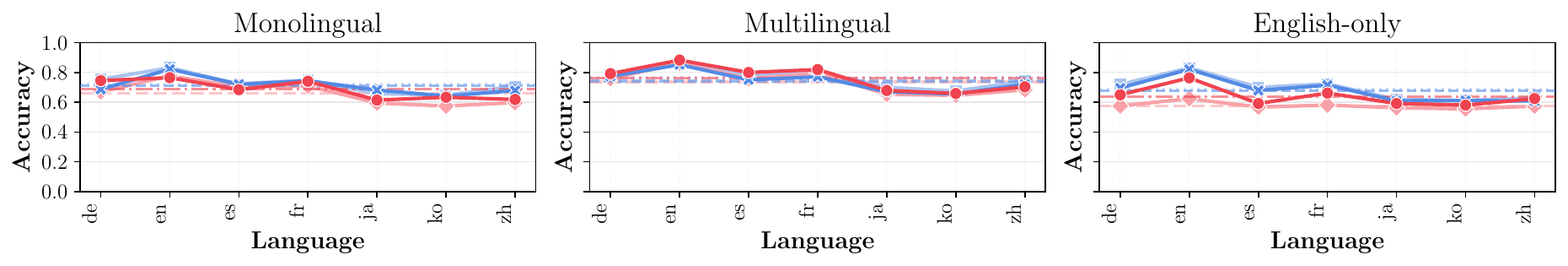}
        \caption{PAWS-X.}
        \label{fig:downstream_pawsx}
    \end{subfigure}

    \caption{Per-language fine-tuning results for (a) XNLI and (b) PAWS-X showing \textcolor{text}{\textbf{Text Opt}}, \textcolor{ipa}{\textbf{IPA Opt}}, \textcolor{textsub}{\textbf{Text Subopt}}, and \textcolor{ipasub}{\textbf{IPA Subopt}} models. We show results for three different fine-tuning strategies (monolingual, multilingual, and English-only). \textbf{Takeaway:} On average, \textcolor{ipa}{\textbf{IPA}} models are on-par with \textcolor{text}{\textbf{Text}} on both datasets, with some minor variation across fine-tuning strategies.}
    \label{fig:downstream}
\end{figure*}

\begin{figure}[htbp]
    \centering
    
    \begin{subfigure}{.45\textwidth}
        \centering
        \includegraphics[width=\linewidth]{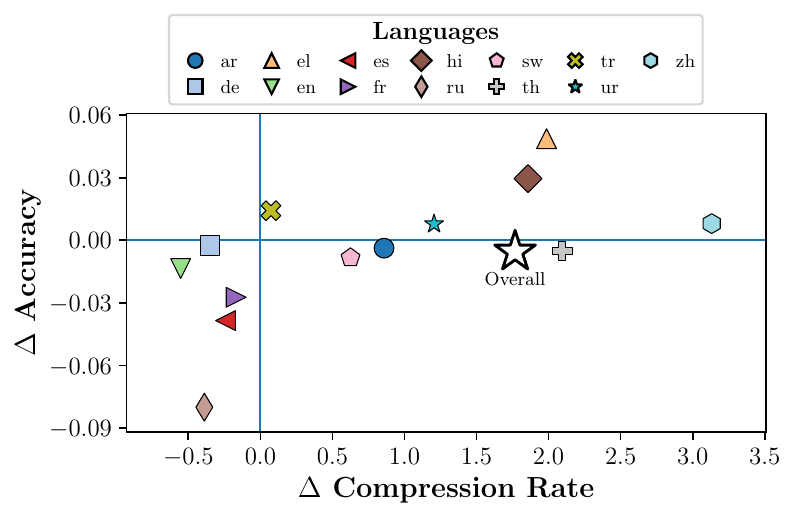}
        \caption{XNLI.}
        \label{fig:compression_xnli}
    \end{subfigure}
    
    \begin{subfigure}{.45\textwidth}
        \centering
        \includegraphics[width=\linewidth]{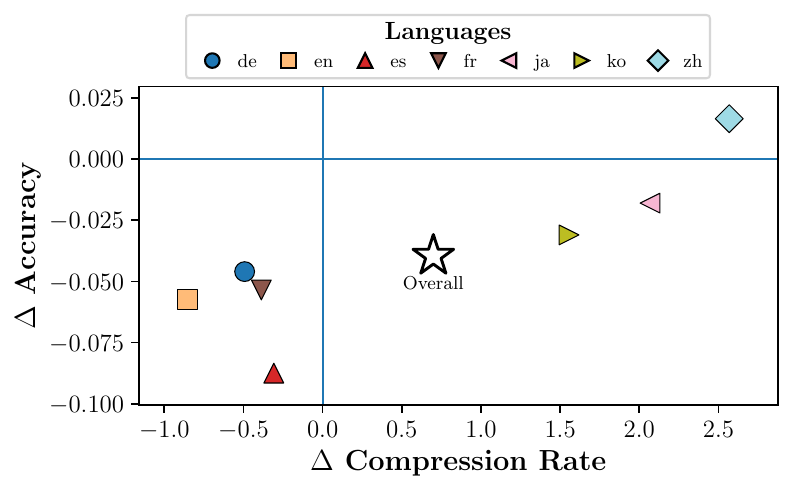}
        \caption{PAWS-X.}
        \label{fig:compression_pawsx}
    \end{subfigure}

    \caption{Accuracy--compression trade-off of \textcolor{ipa}{\textbf{IPA Opt}} relative to \textcolor{text}{\textbf{Text Opt}} on (a) XNLI and (b) PAWS-X. Each point is a language with $\Delta\mathrm{Acc}=\mathrm{Acc}_{\textcolor{ipa}{\mathrm{IPA\,Opt}}}-\mathrm{Acc}_{\textcolor{text}{\mathrm{Text\,Opt}}}$ versus $\Delta\mathrm{CR}=\mathrm{CR}_{\textcolor{ipa}{\mathrm{IPA\,Opt}}}-\mathrm{CR}_{\textcolor{text}{\mathrm{Text\,Opt}}}$; the large star denotes the macro-average across languages. Points further right indicate better compression and points higher indicate better accuracy. \textbf{Takeaway:} \textcolor{ipa}{\textbf{IPA}} models improve overall compression, resulting in reduced inference cost.}
    \label{fig:compression}
\end{figure}

\subsection{Intrinsic tokenization quality across languages and configurations}
\label{sec:results-intrinsic}

We begin with two complementary diagnostic views: the configuration-level heatmaps (Fig.~\ref{fig:heatmap}), which compare intrinsic performance across all tokenizer settings for \textcolor{text}{\textbf{Text}} vs.\ \textcolor{ipa}{\textbf{IPA}}, and a detailed per-language evaluation (Fig.~\ref{fig:intrinsic_per_lang}) of the optimal tokenizers (with mean statistics for suboptimal ones).

\paragraph{IPA consistently improves tokenization quality.} Under their respective optimal settings, IPA improves overall tokenization quality across languages compared to Text (Fig.~\ref{fig:intrinsic_per_lang}). Across all tested configurations, IPA records more notable wins than Text (158 vs.\ 118; $\Delta{>}0.1$ in the normalized metric scale), as shown in Fig.~\ref{fig:heatmap}, with especially consistent advantages on compression metrics (WF, PCW, ATL, CR) and cross-lingual equity (TFG).Text, by contrast, consistently achieves higher Rényi entropies, indicating a more even spread of token usage across the vocabulary. This pattern is best interpreted in light of IPA’s compression gains: higher compression suggests that the learned units are, on average, more word-like, which in turn yields a more Zipfian token distribution, concentrating probability mass more strongly on frequent units while producing a longer tail of rare tokens. Rényi entropy inevitably drops in this scenario, but not as a sign of poorer tokenization, but as a consequence of a more coherent subword structure, which is generally a desirable property. Fig.~\ref{fig:metric_impact} further summarizes paired effect sizes $d_z$ per metric, aggregated across languages ($d_z\!>\!0 \Rightarrow$ IPA is better than Text; see Appendix~\ref{app:wr_dz_zbar}), corroborating that IPA’s gains are systematic rather than driven by a single language or setting.

\paragraph{Sensitivity to different configurations.} Independent of representation, several configuration trends are stable across metrics: larger vocabularies generally improve compression and yield lower Zipf deviation (best ZipfD typically at 200k), while R\'enyi entropies show a trade-off for different orders ($H_2$ and $H_\infty$ strongest at small vocabularies (40k) but $H_1$ tending to peak at the largest (200k)). Vocabulary usage metrics show a similar pattern: with VU highest at 40k and TTR at 200k vocabularies. Cross-lingual equity prefers smaller vocabularies under the balanced sampling strategies, but shifts this preference towards larger vocabularies under unbalanced mixtures. We provide a more detailed interpretation of these effects in Appendix~\ref{apdx:config_effects}. Finally, the optimal Text and IPA tokenizers emerge under different configurations, as outlined in Section~\ref{sec:gpt2} and highlighted in Fig.~\ref{fig:heatmap}.

\subsubsection{Metric-level analysis}

\paragraph{IPA brings largest gains in compression and cross-lingual equity.} Figure~\ref{fig:metric_impact} shows that IPA’s largest metric improvements are in compression-related measures (ATL and CR) and cross-lingual equity (TFG), followed by consistent gains in token-frequency distribution (R\'enyi entropies across different orders). Compression gains indicate that IPA achieves more compact segmentations (fewer tokens for comparable content), while improved equity indicates more uniform tokenization quality across languages. In contrast, improvements are smaller for vocabulary utilization (VU) and continued-word behavior (PCW). The smaller improvements in VU are likely a combined result of our very large vocabularies ($200$k) and small test sets (WikiPron and FLORES+). With a large vocabulary, many tokens sit in a long tail that never appears in the evaluation sample. Smaller improvements in PCW compared to other compression metrics suggest that while words are overall split into fewer tokens, the IPA tokenizers still rarely store entire words in their vocabularies.

\subsubsection{Language- and script-level analysis}

\paragraph{Gains concentrate on non-Latin and unseen scripts.} To characterize how different languages are impacted, we use two complementary summaries over intrinsic metrics: \emph{win rate} (WR), the fraction of metrics where IPA outperforms Text for a given language/script, and the \emph{mean z-score} ($\bar{z}$), which summarizes the average standardized magnitude of IPA--Text differences (both described in more detail in Appendix~\ref{app:wr_dz_zbar}). The language- and script-level rankings (Fig.~\ref{fig:per_lang_winrate}, Fig.~\ref{fig:per_script_winrate}) show that the largest improvements concentrate on non-Latin scripts, with the strongest gains on unseen scripts (\textsc{Grek}, \textsc{Ethi}, \textsc{Mymr}, \textsc{Hebr}). Languages in these scripts (e.g., \textsc{el, am, my, he}) achieve $\mathrm{WR}{=}1$, indicating that IPA improves all intrinsic metrics compared to Text. Substantial gains also appear for complex orthographies and segmentation regimes (e.g., \textsc{lo, th, ja} ($\mathrm{WR}{=}1$), and \textsc{zh} ($\mathrm{WR}{=}0.9$)), morphologically rich languages (e.g., \textsc{fi} ($\mathrm{WR}{=}0.9$) and \textsc{tr} ($\mathrm{WR}{=}0.72$)), as well as low-resource languages (e.g., \textsc{lo} ($\mathrm{WR}{=}1$) and \textsc{sw} ($\mathrm{WR}{=}0.81$)). These results uncover a particular fragility of Text tokenizers: when a language or script is underrepresented or completely unseen during tokenizer training, subword algorithms often split words into individual characters or even bytes, leading to fragmented segmentation. In contrast, IPA's shared symbol space across scripts better preserves tokenization quality in these situations.

\paragraph{Only a small set of high-resource Latin languages shows mild degradation.} A small subset of languages---\textsc{en, de, fr, es, pl, ru}---shows $\mathrm{WR}{<}0.5$ (Fig.~\ref{fig:per_lang_winrate}), meaning Text wins on a majority of intrinsic metrics for these cases. These languages are predominantly high-resource and written in Latin script, which is best represented during tokenizer training, where Text tokenizers already perform well\footnote{One exception is Russian, which is written in Cyrillic script, but is still one of the better represented languages in our training data.}. However, these degradations have relatively small magnitudes ($\bar{z}{<}0$) compared to the improvements across other languages.

\subsection{Downstream task performance and efficiency}
\label{sec:results-downstream}

\paragraph{IPA models match downstream accuracy while reducing inference cost.} Figure~\ref{fig:downstream} shows per-language accuracies for Text and IPA tokenizers under both \emph{Opt} and \emph{Subopt} configurations, across the three fine-tuning regimes described in Section~\ref{sec:gpt2}. Overall, IPA models achieve mean accuracies comparable to those of Text models on both datasets. On PAWS-X, performance varies modestly across fine-tuning settings: IPA slightly outperforms Text under multilingual fine-tuning, but trails it in the English-only regime. Notably, IPA models outperform Text models on Greek (\textsc{el}), the only language in either benchmark absent from the tokenizer training data. We also observe a small set of languages (\textsc{th}, \textsc{zh}), for which IPA models consistently lag slightly behind Text models; one possible explanation is somewhat lower grapheme-to-phoneme quality for these languages, though a detailed per-language analysis is left for future work. Despite this near-parity in accuracy, the accuracy--compression trade-off in Fig.~\ref{fig:compression} shows that IPA improves overall compression on both benchmarks. In practice, this means fewer tokens per input and therefore lower inference cost. Consistent with the intrinsic results, the same set of high-resource Latin languages (\textsc{en, de, fr, es}) + \textsc{ru} see mild compression regressions, which are outweighed by more substantial improvements on other languages, resulting in a favorable accuracy--efficiency trade-off: near-par downstream performance with reduced token counts, translating to lower compute and latency at inference.

\section{Conclusion}
In this work, we show that using IPA as an input representation results in improved and more equitable multilingual tokenization quality. Across 24 languages and 14 scripts, the optimal IPA tokenizer consistently outperforms the optimal Text tokenizer on our suite of 10 intrinsic metrics, with largest gains in compression and cross-lingual equity, and most pronounced improvements for non-Latin and unseen scripts. Models pretrained on IPA match text-based models on downstream performance (XNLI, PAWS-X), while lowering overall inference cost due to improved compression. This establishes IPA as a promising direction for reducing tokenization-driven disparities in MLMs.

\section{Limitations}
\paragraph{Dependence on G2P quality.} Our approach relies on grapheme-to-phoneme (G2P) tools to obtain IPA transcriptions at scale. In practice, G2P quality is not even across all languages, and rule-based or deterministic tools such as Epitran cannot handle pronunciation variation (e.g., British English vs. American English) or homographs, where identical spellings correspond to different pronunciations (e.g., live as a verb vs. adjective). As multilingual G2P systems improve and become more robust, IPA-based pipelines should become increasingly practical for large-scale pretraining.

\paragraph{Converting from IPA back to Text.} Autoregressive models trained purely on IPA cannot directly generate standard orthographic text. IPA is a lossy encoding, and not always is there an unambiguous mapping back to the original writing system, making straightforward decoding a challenge. Moreover, to the best of our knowledge, there currently exist no publicly available automatic multilingual phoneme-to-grapheme (P2G) conversion tools. Since many real-world uses of generative language models require fluent orthographic output, future work could focus on developing accurate P2G tools or finding ways to incorporate the IPA-to-text mapping mechanism directly into the model architecture.

\paragraph{Generalizability.} While we cover a diverse set of languages and vary tokenization and data-sampling configurations to improve generalizability of our findings, our pretrained GPT-2 models (240M parameters) are rather small by current standards. It therefore remains an open question whether the same trade-offs hold at larger scales. In addition, we evaluate downstream transfer on only two tasks and report a single run per model. Stronger evidence would come from broader downstream evaluation and reporting average results over multiple runs to improve robustness.

\paragraph{Representation choices and scope of analysis.} In this work, we focus on IPA transcriptions and do not consider romanization as an alternative input representation. Romanization converts text from its original script into a Latin-script representation, offering similar benefits as IPA in terms of reducing script variation and increasing cross-lingual consistency. In addition, our analysis primarily targets the tokenization stage. While we do provide a set of downstream results, we do not provide a detailed per-language analysis. Future work could therefore compare orthographic text, IPA, and romanization more systematically across the stages of pre-training and fine-tuning, and analyze the effects of these different input representations on downstream performance more thoroughly.

\section{Acknowledgements}

This research was funded by the European Union (ERC, CulturAL, 101171968). Views and opinions expressed are however those of the author(s) only and do not necessarily reflect those of the European Union or the European Research Council. Neither the European Union nor the granting authority can be held responsible for them. 

\erceulogo

\bibliography{anthology-1, anthology-2, custom}

\appendix

\section{International Phonetic Alphabet (IPA)}\label{apdx:ipa}

The International Phonetic Alphabet (IPA) is a system of phonetic notation developed and maintained by the International Phonetic Association to provide a standardized, language-independent representation of the sounds of spoken language. Originally created in the late 19th century, the IPA has undergone multiple refinements over the years, with the most recent official chart standardized in 2020\footnote{\url{https://www.internationalphoneticassociation.org/content/full-ipa-chart}}.

The IPA uses a set of dedicated symbols\footnote{The exact number varies by source, depending on IPA variants and criteria for what counts as a single character, but typically ranges between 100 and 200 unique symbols.} to represent speech sounds, distinguishing four main categories: consonants, vowels, diacritics, and suprasegmentals. Consonant and vowel symbols correspond to specific phonemes and are organized according to articulatory features such as place and manner of articulation for consonants, and tongue height and backness for vowels. Diacritics are used to modify base symbols to indicate finer phonetic details such as nasalization, aspiration, or devoicing. Suprasegmentals, such as stress, length, intonation, and tone, are represented with additional symbols and markings to capture prosodic aspects of speech.

\begin{table*}[t]
  \centering
  \small
  \setlength{\tabcolsep}{6pt}      
  \renewcommand{\arraystretch}{1.15}
  \begin{tabular}{@{}lll@{}}
    \toprule
    \textbf{Word-level} & \textbf{Subword-level} & \textbf{Character-/ byte-level} \\
    \midrule
    Moses\textsuperscript{*} \citep{koehn2007moses}             & Morfessor\textsuperscript{\textdagger} \citep{creutz2005morfessor}         & Charformer \citep{tay2021charformer}     \\
    NLTK\textsuperscript{*}  \citep{bird-2009-nltk}                 & WordPiece\textsuperscript{\textdagger} \citep{schuster2012wordpiece}             & ByT5 \citep{xue2022byt5} \\
    spaCy\textsuperscript{*} \citep{Honnibal_spaCy_Industrial-strength_Natural_2020}    & BPE\textsuperscript{\textdagger} \citep{sennrich-etal-2016-bpe}       & CANINE \citep{clark2022canine} \\
    \multicolumn{1}{c}{}                 & UnigramLM\textsuperscript{\textdagger} \citep{kudo2018unigram}           & MegaByte \citep{yu2023megabyte} \\
    \multicolumn{1}{c}{}                 & \multicolumn{1}{c}{}               & SpaceByte \citep{slagle2024spacebyte}  \\
    \multicolumn{1}{c}{}                 & \multicolumn{1}{c}{}               & BLT \citep{pagnoni2024blt} \\
    \multicolumn{1}{c}{}                 & \multicolumn{1}{c}{}               & MrT5 \citep{kallini2024mrt5} \\
    \bottomrule
  \end{tabular}
  \caption[Taxonomy of prominent tokenization methods]{Representative tokenizers grouped by granularity. Asterisk symbol (*) marks rule-based tokenizers, and dagger symbol (\textdagger) data-driven methods. Character-/byte-level column names tokenizer-free models that operate on raw characters/bytes using different approaches.}
  \label{tab:token-taxonomy}
\end{table*}

Notably, there are two common modes of IPA transcriptions---broad and narrow---which differ in the level of phonetic detail. Broad transcriptions are \textit{phonemic}, in that they only include contrastive sounds needed to distinguish meaning, while narrow transcriptions are \textit{allophonic}, encoding more fine-grained information about phonetic realization and variation. For example, a broad transcription might not differentiate between the sound corresponding to the letter ``l'' in ``leaf'' and ``pool'' (/lif/ and /pul/), while a narrow transcription would (\textipa{[li:f]} and \textipa{[pu:ɫ]})\footnote{It would also not differentiate vowel length (indicated by ``\textipa{:}''), unless this feature is contrastive (i.e. changes word meaning) in the particular language.}\textsuperscript{,}\footnote{Here we adopt the notation from the official IPA Handbook of enclosing phonemes within forward slashes, i.e. ``/\ldots/'', and phones within square brackets, i.e. ``[\ldots]''}. For more information about IPA, one can consult the detailed handbook provided by the \citet{ipa1999}.

\section{Tokenization}\label{apdx:tok}

Tokenization---the process of dividing a sequence of text\footnote{While we focus on text tokenization here, the concept applies more broadly to different types of sequential data e.g., audio stream in speech models \citep{zhang2023speechtokenizer} or image pixels in computer vision models \citep{qian2022imagetok}.} into smaller, discrete units (i.e. \textit{tokens})---has been an active area of research in recent years \citep[\textit{inter alia}]{hou2023effects, domingo2019much, rajaraman2024toward}. It comes as the first component in the process of training an LLM, and has been shown to have a significant impact on model performance and generalization, especially in multilingual settings \citep{chelombitko2024qtok}.

Tokenization methods can be systematically classified along two axes: (1) the approach to determining token boundaries—\textit{rule-based} versus \textit{data-driven}; and (2) the granularity of the resulting tokens—\textit{word-level}, \textit{subword-level}, and \textit{character- or byte-level} \citep{gastaldi2024foundations}. Rule-based tokenization relies on predefined, usually linguistically motivated heuristics, while data-driven approaches learn token boundaries directly from data, typically optimizing for frequency, likelihood, or other corpus-derived statistics. Regarding granularity, in word-level tokenization, each token corresponds to a complete word form. Subword-level tokenization segments text into units smaller than or equal to full words. Character- and byte-level methods treat each character or byte as a separate token, respectively. Table~\ref{tab:token-taxonomy} categorizes some notable tokenizers along these axes.

Early NLP systems relied predominantly on rule-based, word-level tokenization. However, this approach encounters significant limitations such as rapid vocabulary growth, frequent out-of-vocabulary (OOV) issues, and necessity for language-specific hand-crafted rules that often require linguistic expertise \citep{mielke2021between}. Data-driven, subword-level tokenizers emerged to address these issues, offering several key advantages. They significantly reduce vocabulary size, improve generalization, and efficiently handle OOV terms by supporting open-vocabulary modeling \citep{gastaldi2024foundations}. However, subword tokenization is not without drawbacks. Most notably for our discussion, these methods might neglect meaningful linguistic structures \citep{bostrom2020byte, beinborn2023analyzing} and often unfairly allocate vocabulary space, disproportionately favoring high-resource languages in the training data \citep{qin2025survey, mielke2021between, rust2020good}. Recently, fully character- and byte-level methods (also collectively known as ``tokenization-free'' methods) gained attention as alternatives. They entirely avoid vocabulary constraints, providing universal coverage and script independence, alongside robustness to noise \citep{xue2022byt5}. Yet, these methods also face certain limitations. Firstly, they significantly increase computational cost due to substantially longer input sequences \citep{kallini2024mrt5}. Secondly, by operating on raw bytes, these models inevitably inherit a non-linguistically motivated Unicode\footnote{\url{https://home.unicode.org/}}-based scheme. In this scheme, every Latin character is encoded in a single byte, whereas characters in many other scripts (e.g.\ Devanagari, Cyrillic, Han) require two to four bytes per character. Consequently, a Hindi or Chinese sentence of the same length (in terms of characters) as an English one is broken into far more byte‑tokens, inflating sequence length and compute and giving those languages a systematic efficiency disadvantage \citep{mielke2021between}. Current research aims to resolve the challenges highlighted above; some notable examples have been highlighted in Section~\ref{sec:rel_work}.

\section{Data and languages}\label{apdx:data_and_langs}

For convenience, we provide a mapping between all of the languages considered in this study and their ISO codes in Table~\ref{tab:apd_languages}.

\subsection{Mapping of languages and scripts to their ISO codes}\label{apdx:iso_mappings}

\begin{table}[ht]
\small
\centering
\begin{tabular}{llll}
\toprule
\textbf{Language} & \textbf{ISO 639-1} & \textbf{Script} & \textbf{ISO 15924} \\
\midrule
Arabic         & ar  & Arabic            & Arab \\
Amharic        & am  & Ethiopic          & Ethi \\
Burmese        & my  & Myanmar           & Mymr \\
Chinese        & zh  & Han               & Hani \\
English        & en  & Latin             & Latn \\
Finnish        & fi  & Latin             & Latn \\
French         & fr  & Latin             & Latn \\
German         & de  & Latin             & Latn \\
Greek\textsuperscript{\textdagger}           & el  & Greek             & Grek \\
Hebrew\textsuperscript{\textdagger}         & he  & Hebrew            & Hebr \\
Hindi          & hi  & Devanagari        & Deva \\
Italian        & it  & Latin             & Latn \\
Japanese\textsuperscript{*}        & ja  & Japanese & Jpan \\
Korean         & ko  & Hangul            & Hang \\
Lao            & lo  & Lao               & Laoo \\
Persian        & fa  & Arabic            & Arab \\
Polish         & pl  & Latin             & Latn \\
Russian        & ru  & Cyrillic          & Cyrl \\
Serbian\textsuperscript{*}        & sr  & Latin/Cyrillic    & Latn/Cyrl \\
Spanish        & es  & Latin             & Latn \\
Swahili        & sw  & Latin             & Latn \\
Thai           & th  & Thai              & Thai \\
Turkish        & tr  & Latin             & Latn \\
Urdu           & ur  & Arabic            & Arab \\
\bottomrule
\end{tabular}
\caption{Languages used in this work. Each language is followed by its two-letter ISO 639-1 code, its script, and the script’s four-letter ISO 15924 code. Languages marked with an asterisk (*) use multiple scripts, while languages not supported by Epitran are marked with a dagger symbol (\textdagger).}
\label{tab:apd_languages}
\end{table}

\subsection{Data sampling strategies}\label{apdx:sampling_strategies}

\begin{figure*}[t]
  \centering
  \includegraphics[width=2\columnwidth]{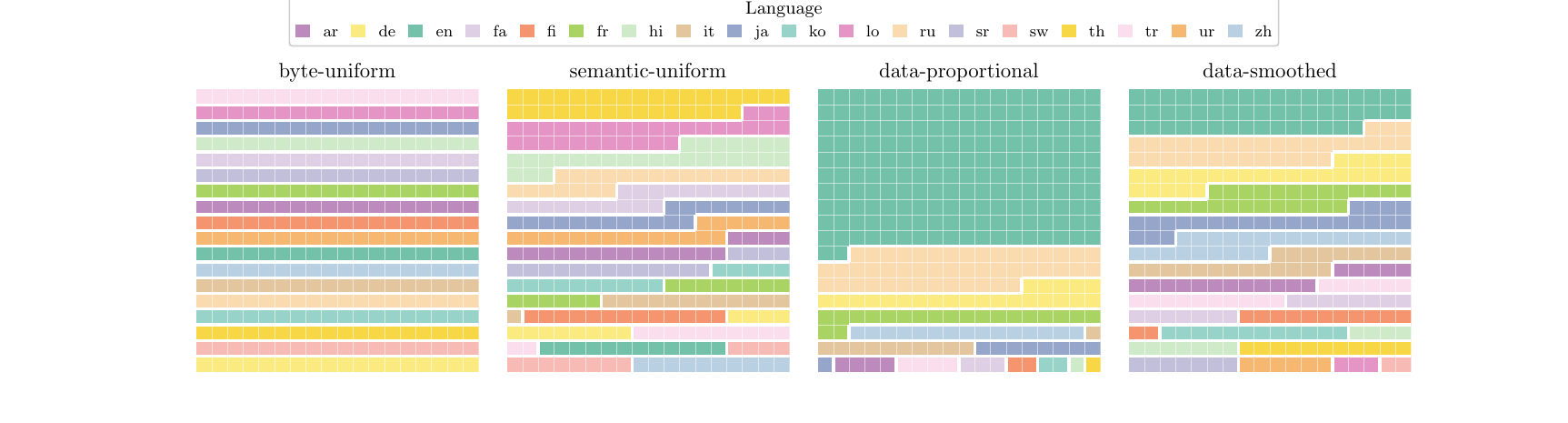}
  \caption{Resulting distribution of languages in the training data using four different sampling strategies. Each square represents $\approx0.3\%$ of the training data.}
  \label{fig:lang_dist}
\end{figure*}

We visualize the resulting language distributions of the four different data sampling strategies in Fig.~\ref{fig:lang_dist}.

\subsection{WikiPron details}\label{apdx:wikipron}

Table~\ref{tab:wikipron-stats} shows WikiPron dataset details by language on the sample that we use in this work.

\begin{table*}[ht]
\centering
\footnotesize
\begin{tabular}{@{}l l r r c c@{}}
\toprule
\multicolumn{6}{c}{\textbf{Seen Languages (18)}} \\ 
\midrule
\textbf{Language} & \textbf{ISO} & \textbf{Original} & \textbf{After Removal} & \textbf{Scope} & \textbf{Filtered} \\
\midrule
Arabic            & ar &  2\,252   &  1\,596   & Broad  & No \\
German            & de & 47\,779   & 42\,914   & Broad  & Yes \\
English           & en & 81\,576   & 69\,639   & Broad  & Yes \\
Persian           & fa & 34\,033   &  7\,357   & Narrow  & No \\
Finnish           & fi &158\,880   &156\,923   & Broad  & No \\
French            & fr & 80\,690   & 71\,229   & Broad  & Yes \\
Hindi             & hi & 24\,640   & 15\,867   & Broad  & Yes \\
Italian           & it & 79\,865   & 73\,136   & Broad  & Yes \\
Japanese*          & ja & 32\,749   & 29\,092   & Narrow  & Yes \\
Korean            & ko & 22\,072   & 21\,383   & Narrow  & Yes \\
Lao               & lo &  4\,180   &  2\,085   & Narrow  & No \\
Russian           & ru &411\,651   &395\,845   & Narrow  & No \\
Serbian*           & sr & 46\,991   & 45\,553   & Broad  & Yes \\
Swahili           & sw &    110    &    106    & Broad  & No \\
Thai              & th & 16\,689   & 11\,589   & Broad  & No \\
Turkish           & tr &  7\,266   &  6\,933   & Broad  & No \\
Urdu              & ur &  4\,493   &  3\,637   & Broad  & No \\
Chinese           & zh &158\,873   &126\,804   & Broad  & No \\
\midrule
\textbf{Total}    &---&\textbf{1\,214\,798}&\textbf{1\,081\,688}&       &     \\
\midrule
\multicolumn{6}{c}{\textbf{Unseen Languages (6)}} \\
\midrule
\textbf{Language} & \textbf{ISO} & \textbf{Original} & \textbf{After Removal} & \textbf{Scope} & \textbf{Filtered} \\
\midrule
Amharic           & am &    378    &    371    & Broad & No \\
Greek     & el & 14\,825   & 14\,825   & Broad & Yes \\
Spanish           & es & 99\,043   & 98\,787   & Broad & Yes \\
Hebrew            & he &  1\,957   &  1\,945   & Broad & No \\
Burmese           & my &  6\,062   &  4\,486   & Broad & Yes \\
Polish            & pl &132\,558   &115\,865   & Broad & No \\
\midrule
\textbf{Total}    &---&\textbf{256\,608}  &\textbf{238\,057}  &       &     \\
\bottomrule
\end{tabular}
\caption{WikiPron details by language. ``Original'' represents the total number of words in the original dataset; ``After Removal'' shows the number of remaining words after deduplication. ``Scope'' and ``Filtered'' columns direct to the specific WikiPron file that was used per language (we always selected \textit{broad} and \textit{filtered} versions, unless unavailable). For languages marked with an asterisk (*), we include different scripts (Katakana and Hiragana for Japanese; Latin and Cyrillic for Serbian).}
\label{tab:wikipron-stats}
\end{table*}

\subsection{CulturaX post-processing}
\label{app:culturax_cleaning}

We apply a lightweight, line-level cleaning step to CulturaX before writing the sampled text to disk. We remove URLs matching \texttt{http(s)://\dots} or \texttt{www.\dots}; remove any remaining tag-like spans of the form \texttt{<\dots>}; remove emoji characters; filter the text to retain only Unicode letters (\texttt{L}), numbers (\texttt{N}), punctuation (\texttt{P}), and standard spaces (\texttt{Zs}), deleting all other characters; and finally normalize whitespace by collapsing consecutive whitespace to a single space and stripping leading/trailing spaces. After this pipeline, we discard empty lines and keep non-empty cleaned lines for sampling.

\section{Epitran usage details}\label{apdx:epitran}

A typical usage involves instantiating an \texttt{Epitran} object with a language–script code (e.g., \\\texttt{epitran.Epitran('eng-Latn')} for English) and then calling its \texttt{transliterate()} method, which returns the IPA representation. Internally, Epitran relies on language‑specific mapping files and optional pre- and post-processing rules. An alternative utility is the \texttt{Backoff} class, which allows cascading multiple language–script mappings (for example, to handle mixed-script inputs by falling back from one transliterator to another). However, the \texttt{Backoff} class currently does not support the full range of text pre- and post-processing routines applied by the standard \texttt{transliterate()}, leading us to adopt a different transliteration approach. In this section, we describe this approach, as well as other modifications that we made to this tool.

\paragraph{Bug fixes and mapping extensions.} We corrected identified bugs in the Serbian mapping and added several missing mappings for Urdu, Arabic, Persian, Hindi, Finnish, and Turkish. These additions were based on empirically observed character sequences occurring above 1\% frequency in our training data and consulted via well-documented Wikipedia pages of the respective languages. The exact modifications we made per aforementioned language are accessible in our codebase.

\begin{figure*}[t]
\centering
\begin{subcaptionbox}{Chinese}[0.3\textwidth]
  {\includegraphics[width=\linewidth]{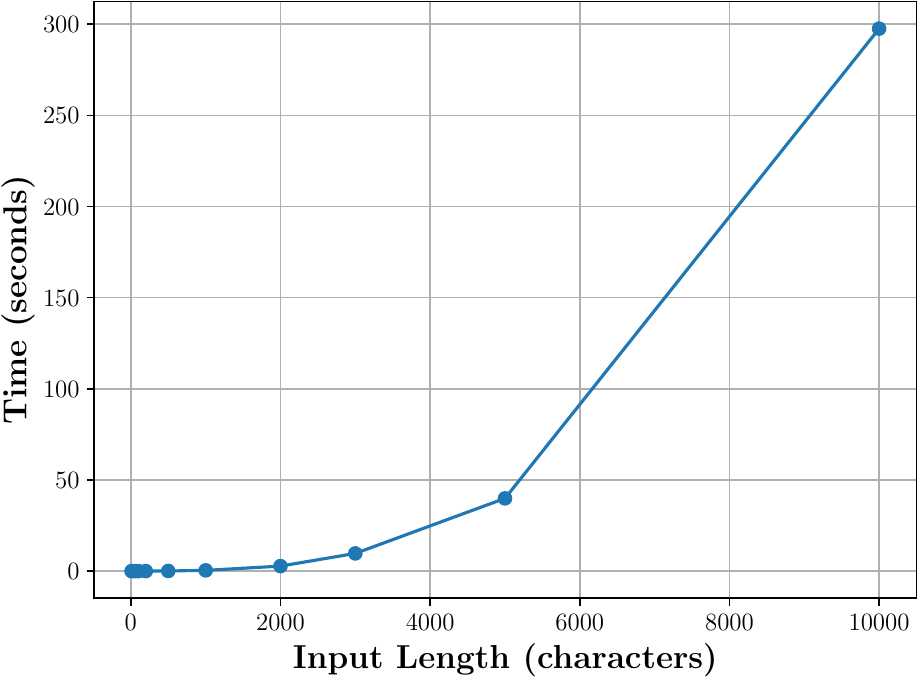}}
\end{subcaptionbox}
\hspace{0.05\textwidth}
\begin{subcaptionbox}{English}[0.3\textwidth]
  {\includegraphics[width=\linewidth]{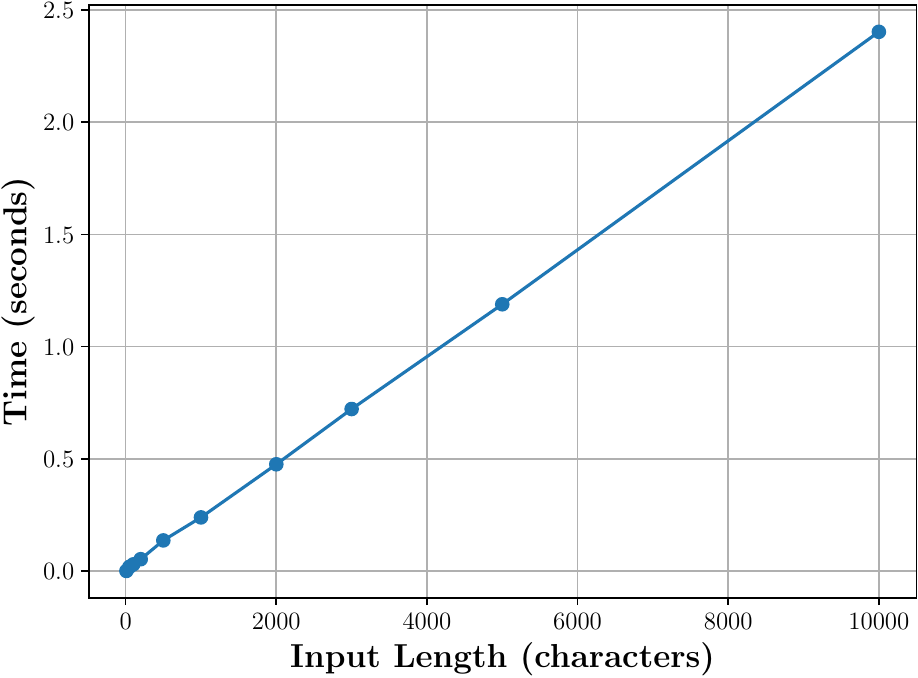}}
\end{subcaptionbox}
\caption[Epitran: Runtime vs. Input length efficiency trade-off]{Runtime vs. Input length efficiency trade-off for (a) Chinese and (b) English.}
\label{fig:time_length_tradeoff}
\end{figure*}

\begin{figure*}[t]
\centering
\begin{subcaptionbox}{Chinese}[0.3\textwidth]
  {\includegraphics[width=\linewidth]{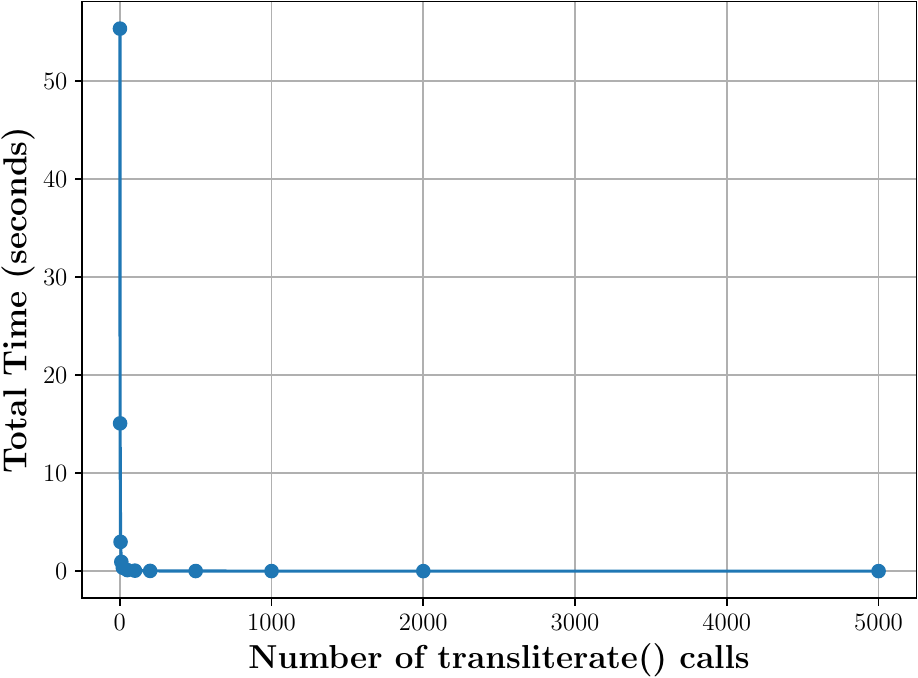}}
\end{subcaptionbox}
\hspace{0.05\textwidth}
\begin{subcaptionbox}{English}[0.3\textwidth]
  {\includegraphics[width=\linewidth]{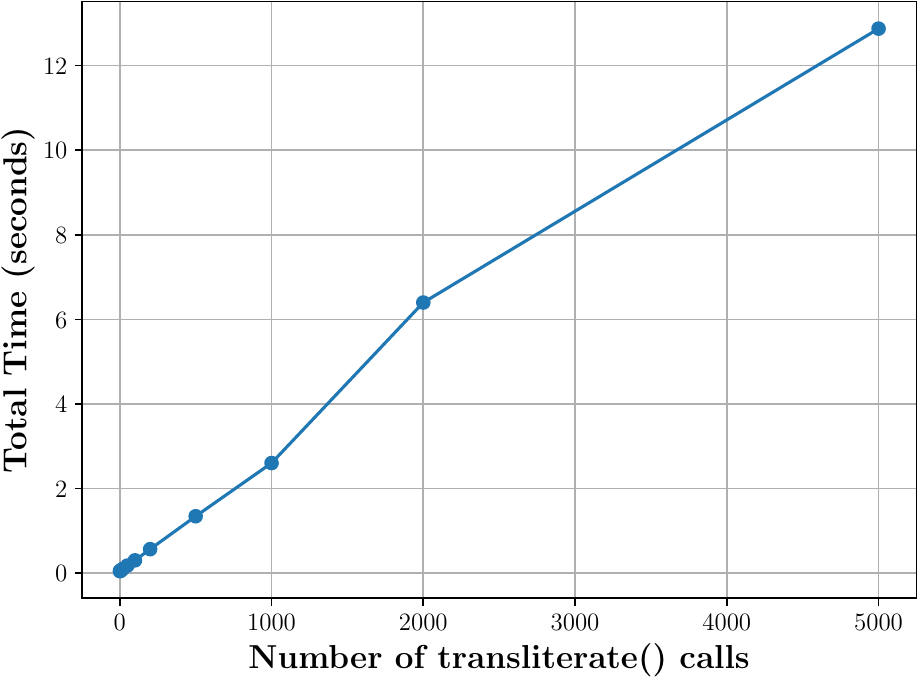}}
\end{subcaptionbox}
\caption[Epitran: Runtime vs. Number of \texttt{transliterate()} calls efficiency trade-off]{Runtime vs. Number of \texttt{transliterate()} calls efficiency trade-off for (a) Chinese and (b) English.}
\label{fig:time_calls_tradeoff}
\end{figure*}

\paragraph{Efficient processing for Chinese and Japanese.} While most languages were processed by applying \texttt{transliterate()} \textit{once} per full document, this approach proved inefficient for Chinese and Japanese due to their special treatment by Epitran under the hood, which resulted in a non-linear runtime increase with increased input length (see Figure~\ref{fig:time_length_tradeoff}). To address this, we split documents into smaller segments along Chinese/Japanese specific punctuation, ensuring word boundaries were preserved. This chunking preserved performance while maintaining accuracy. In contrast, this segmentation approach results in a notable overhead for other languages (see Figure~\ref{fig:time_calls_tradeoff}), so it was only applied to Chinese and Japanese.

\paragraph{Sequential mapping for multiple variants.} For languages supported by multiple mapping files we employed a specific strategy: we applied the first mapping to the full text, then identified characters still untransliterated (i.e., non-IPA), and selectively applied subsequent mappings only to those residual characters. This procedure ensured each script variant was handled thoroughly, while avoiding unnecessary repeated processing.

\paragraph{Post-conversion cleanup.} After IPA conversion, any remaining characters not recognized or mapped by Epitran (primarily external script tokens included in CulturaX data for the particular language, but also, at times, characters not supported by Epitran) were removed. Across languages, this accounted for an average of only 0.3\% of tokens per language, a removal rate negligible in volume yet important for ensuring consistency and purity of the phonological representation space.

\section{G2P conversion quality}\label{apdx:g2p_quality}

We estimated the quality of Epitran grapheme-to-phoneme transcriptions by comparing each predicted IPA form against gold pronunciations from WikiPron using normalized Feature Error Rate (nFER). To motivate this choice, we first briefly relate nFER to the more standard Phoneme Error Rate (PER) and to Feature Error Rate (FER), from which nFER is derived.

\paragraph{Phoneme Error Rate (PER).} Let $w$ be a word, let $\hat{\mathbf{y}}=(\hat{y}_1,\dots,\hat{y}_m)$ denote its predicted IPA transcription, and let $\mathcal{G}(w)=\{\mathbf{y}^{(1)},\dots,\mathbf{y}^{(K)}\}$ be the set of gold IPA variants in WikiPron. PER is the standard phone-level Levenshtein distance \citep{levenshtein1966binary} between a prediction $\hat{\mathbf{y}}$ and a gold transcription $\mathbf{y}$, normalized by the number of phones in the reference:
\begin{equation}
\mathrm{PER}(\hat{\mathbf{y}},\mathbf{y})
=
\frac{\mathrm{ED}_{\mathrm{phone}}(\hat{\mathbf{y}},\mathbf{y})}{|\mathbf{y}|},
\end{equation}
where $\mathrm{ED}_{\mathrm{phone}}$ uses insertion, deletion, and substitution costs of $1$. PER therefore treats all phone substitutions equally, regardless of whether the mismatched phones are phonetically very similar or very different.

\paragraph{Feature Error Rate (FER).} FER replaces this binary substitution cost with a feature-based cost derived from PanPhon \citep{mortensen2016panphon} articulatory feature vectors. Let $\phi(p)\in\{-1,0,+1\}^F$ denote the feature vector for phone $p$, where $F$ is the number of articulatory features. The substitution cost between phones $p$ and $q$ is
\begin{equation}
c_{\mathrm{sub}}(p,q)
=
\frac{1}{F}\sum_{f=1}^{F}\frac{|\phi_f(p)-\phi_f(q)|}{2}.
\end{equation}
This assigns cost $0$ to identical phones, smaller penalties when one feature value is unspecified, and larger penalties when phones differ on many specified features. Insertions and deletions are defined analogously from the feature vector of the inserted or deleted phone, yielding a feature-based edit distance $\mathrm{ED}_{\mathrm{feat}}(\hat{\mathbf{y}},\mathbf{y})$. FER is then
\begin{equation}
\mathrm{FER}(\hat{\mathbf{y}},\mathbf{y})
=
\frac{\mathrm{ED}_{\mathrm{feat}}(\hat{\mathbf{y}},\mathbf{y})}{|\mathbf{y}|}.
\end{equation}

\paragraph{Normalized Feature Error Rate (nFER).} This metric uses the same feature-based edit distance as FER but normalizes by the length of the longer transcription:
\begin{equation}
\mathrm{nFER}(\hat{\mathbf{y}},\mathbf{y})
=
\frac{\mathrm{ED}_{\mathrm{feat}}(\hat{\mathbf{y}},\mathbf{y})}
{\max\!\bigl(|\hat{\mathbf{y}}|,|\mathbf{y}|\bigr)}.
\end{equation}
This choice makes the score less sensitive to length mismatches between the predicted and gold transcriptions. It is particularly useful here because Epitran and WikiPron sometimes differ slightly in IPA conventions while remaining phonetically close; unlike PER, nFER gives lower penalty to such near-matches by measuring disagreement at the level of articulatory features.

Because WikiPron may provide multiple acceptable pronunciations for a word, we score each prediction against all listed variants and retain the best match:
\begin{equation}
\mathrm{nFER}(w)
=
\min_{\mathbf{y}\in\mathcal{G}(w)}
\mathrm{nFER}(\hat{\mathbf{y}},\mathbf{y}).
\end{equation}
We then report the mean over all evaluated words per language:
\begin{equation}
\mathrm{nFER}_{\mathrm{avg}}
=
\frac{1}{|\mathcal{W}|}\sum_{w\in\mathcal{W}}\mathrm{nFER}(w).
\end{equation}
The results are shown in Table~\ref{apdx:nfer}.

\begin{table}[t]
\centering
\small
\begin{tabular}{cS[table-format=2.2]}
\toprule
\textbf{Language} & {\textbf{nFER (\%)}} \\
\midrule
\textsc{am} & 4.00 \\
\textsc{ar} & 24.83 \\
\textsc{de} & 6.77 \\
\textsc{en} & 6.18 \\
\textsc{es} & 0.35 \\
\textsc{fa} & 19.15 \\
\textsc{fi} & 0.88 \\
\textsc{fr} & 7.60 \\
\textsc{hi} & 5.22 \\
\textsc{it} & 6.13 \\
\textsc{ja} & 1.33 \\
\textsc{ko} & 3.24 \\
\textsc{lo} & 22.58 \\
\textsc{my} & 9.85 \\
\textsc{pl} & 1.14 \\
\textsc{ru} & 5.08 \\
\textsc{sr} & 0.87 \\
\textsc{sw} & 5.60 \\
\textsc{th} & 21.76 \\
\textsc{tr} & 1.39 \\
\textsc{ur} & 16.55 \\
\textsc{zh} & 26.49 \\
\bottomrule
\end{tabular}
\caption{Epitran G2P quality measured as average normalized Feature Error Rate (nFER, \%) on a WikiPron sample of 2{,}000 words per language. Lower is better.}
\label{apdx:nfer}
\end{table}

\paragraph{Many-to-one collision rates.} As we noted in our Limitations section, the mapping between graphemes and phonemes can sometimes be ambiguous. On the one hand, a single orthographic string can be mapped to multiple valid strings in IPA (e.g., English homographs "live" -- as an adjective vs. verb). On the other hand, different orthographic strings can be mapped to the same string in IPA (e.g., English homophones "rain" and "reign"). We call these scenarios \textit{one-to-many} and \textit{many-to-one} collisions, respectively. However, in most languages, these phenomena are negligible, as we show in Table~\ref{apdx:tab:collisions}. While both collision scenarios are non-invertible without context, with the presence of context these can be recovered and learned by the model. Therefore, this is not to be viewed as an inherent downside of IPA, nor a challenge that standard Text models avoid.

\begin{table}[t]
\centering
\small
\begin{tabular}{cc}
\toprule
\textbf{Lang.} & \textbf{Collision rate (\%)} \\
\midrule
\textsc{am} & 1.62 \\
\textsc{ar} & 2.67 \\
\textsc{de} & 2.99 \\
\textsc{el} & 13.68 \\
\textsc{en} & 10.78 \\
\textsc{es} & 5.39 \\
\textsc{fa} & 1.91 \\
\textsc{fi} & 0.50 \\
\textsc{fr} & 42.98 \\
\textsc{he} & 3.44 \\
\textsc{hi} & 5.79 \\
\textsc{it} & 0.75 \\
\textsc{ja} & 0.83 \\
\textsc{ko} & 1.43 \\
\textsc{lo} & 7.47 \\
\textsc{my} & 6.36 \\
\textsc{pl} & 3.17 \\
\textsc{ru} & 12.01 \\
\textsc{sr} & 0.06 \\
\textsc{sw} & 0.00 \\
\textsc{th} & 8.00 \\
\textsc{tr} & 0.97 \\
\textsc{ur} & 2.06 \\
\textsc{zh} & 28.19 \\
\bottomrule
\end{tabular}
\caption{Empirical many-to-one collision rates (\%) per language, estimated on WikiPron. Lower is better.}
\label{apdx:tab:collisions}
\end{table}

\section{Intrinsic tokenization metrics}\label{apdx:intrinsic_metrics}

\subsection{Compression}
Let $W_\ell$ be the set of unique word types for language $\ell$, and let $\tau(\cdot)$ be a tokenizer mapping an input string to a token sequence. Denote the number of produced tokens by $n_\tau(x)=|\tau(x)|$ and the character length of a string by $|x|$ (in the tokenizer's input representation).

\paragraph{Word Fertility (WF).}
Average number of subword tokens per word (lower is better):
\begin{equation}
\mathrm{WF}_\ell \;=\; \frac{1}{|W_\ell|}\sum_{w\in W_\ell} n_\tau(w).
\end{equation}

\paragraph{Proportion of Continued Words (PCW).}
Fraction of word types split into multiple tokens (lower is better):
\begin{equation}
\mathrm{PCW}_\ell \;=\; \frac{1}{|W_\ell|}\sum_{w\in W_\ell}\mathbb{I}\!\left[n_\tau(w)>1\right].
\end{equation}

\paragraph{Average Token Length (ATL).}
Average length of produced tokens, measured in characters of the tokenizer input representation (\textcolor{text}{\textbf{Text}} or \textcolor{ipa}{\textbf{IPA}}); higher indicates longer token segments. Let $\mathcal{T}_\ell=\biguplus_{w\in W_\ell}\tau(w)$ be the multiset of all tokens produced for $W_\ell$, and let $|t|$ be the character length of token $t$ (excluding any boundary marker, e.g.\ leading ``\textbf{\_}''). Then ATL is given as:
\begin{equation}
\mathrm{ATL}_\ell \;=\; \frac{1}{|\mathcal{T}_\ell|}\sum_{t\in \mathcal{T}_\ell} |t|.
\end{equation}

\paragraph{Compression Rate (CR).}
Average number of \emph{raw input characters} per token at the sentence level (higher indicates stronger compression). For a sentence set $S_\ell$:
\begin{equation}
\mathrm{CR}_\ell \;=\; \frac{1}{|S_\ell|}\sum_{s\in S_\ell}\frac{|s|}{n_\tau(s)}.
\end{equation}

\subsection{Token frequency distribution shape}
Let $c(v)$ be the count of token type $v$ on evaluation data for language $\ell$, with total token count $N=\sum_v c(v)$ and empirical probabilities $p(v)=c(v)/N$.

\paragraph{R\'enyi Entropy (RE).}
Measures how evenly token usage is distributed across the vocabulary:
\begin{equation}
H_\alpha \;=\; \frac{1}{1-\alpha}\log \sum_{v} p(v)^\alpha.
\end{equation}
We report $\alpha=1$ (Shannon entropy), $\alpha=2$ (collision entropy), and $\alpha=\infty$ (min-entropy). Higher values indicate a more uniform distribution.

\paragraph{Zipf Deviation (ZipfD).}
Following \citet{lotz-etal-2025-beyond}, we quantify how closely the empirical token rank--frequency distribution follows a Zipfian power law by fitting a linear function to the log--log curve and reporting its mean absolute deviation. Concretely, let tokens be sorted by descending frequency, with rank $r\in\{1,\dots,R\}$ and frequency $f_r$. Define $x_r=\log r$ and $y_r=\log f_r$, and estimate parameters $\beta_0,\beta_1$ via least squares for $f(x)=\beta_0+\beta_1 x$. Zipf deviation is then
\begin{equation}
\mathrm{ZipfD} \;=\; \frac{1}{R}\sum_{r=1}^{R}\left|\beta_0+\beta_1 x_r - y_r\right|.
\end{equation}
Lower ZipfD indicates closer agreement with Zipf's law.

\subsection{Vocabulary usage}
Using the same notation above, let $V$ be the learned tokenizer vocabulary with size $|V|$, and let $U=\{v\in V: c(v)>0\}$ be the set of observed token types.

\paragraph{Vocabulary Utilization (VU).}
Share of the learned vocabulary used at least once on evaluation data (higher is better):
\begin{equation}
\mathrm{VU} \;=\; \frac{|U|}{|V|}.
\end{equation}

\paragraph{Type--Token Ratio (TTR).}
Distinct token types relative to the total number of produced tokens (higher indicates more diverse usage):
\begin{equation}
\mathrm{TTR} \;=\; \frac{|U|}{N}.
\end{equation}

\subsection{Cross-lingual equity}
Let $S^{\parallel}_{\ell}$ be a parallel sentence set paired with English, containing pairs $(s^{(i)}_{\mathrm{en}}, s^{(i)}_{\ell})$.

\paragraph{Tokenization Parity (TP)}
Following the discussion of \citet{petrov2023language}, we compare tokenization length relative to English on parallel sentences:
\begin{equation}
\mathrm{TP}_\ell \;=\; \frac{1}{|S^{\parallel}_{\ell}|}\sum_{i}\frac{n_\tau(s^{(i)}_{\ell})}{n_\tau(s^{(i)}_{\mathrm{en}})}.
\end{equation}
Values closer to $1$ indicate more similar tokenization length relative to English.

\paragraph{Tokenization Fairness Gini (TFG).}
Following \citet{meister_tokenizer_analysis_2025}, we measure inequality in tokenization cost across languages using the (equal-weight) Gini coefficient. Let $\mathcal{L}=\{1,\dots,n\}$ be the set of languages. For each language $\ell\in\mathcal{L}$, define the token cost as
\begin{equation}
c_\ell \;=\; 
\frac{\sum_{s\in S_\ell} n_\tau(s)}
{\sum_{s\in S_\ell} |s|_{\mathrm{bytes}}}
\, ,
\end{equation}
where $n_\tau(s)$ is the number of tokens produced for sentence $s$ and $|s|_{\mathrm{bytes}}$ is its UTF-8 byte length. The mean cost is:
\begin{equation}
\mu \;=\; \frac{1}{n}\sum_{\ell=1}^{n} c_\ell \, .
\end{equation}
TFG is then:
\begin{equation}
\mathrm{TFG} \;=\;
\frac{\sum_{i=1}^{n}\sum_{j=1}^{n}\left|c_i-c_j\right|}
{2n^{2}\mu}
\, .
\end{equation}
Lower TFG indicates more equal byte-normalized token costs across languages.

\begin{figure*}[t]
  \centering
  \begin{subfigure}[t]{0.49\textwidth}
    \centering
    \includegraphics[width=\linewidth]{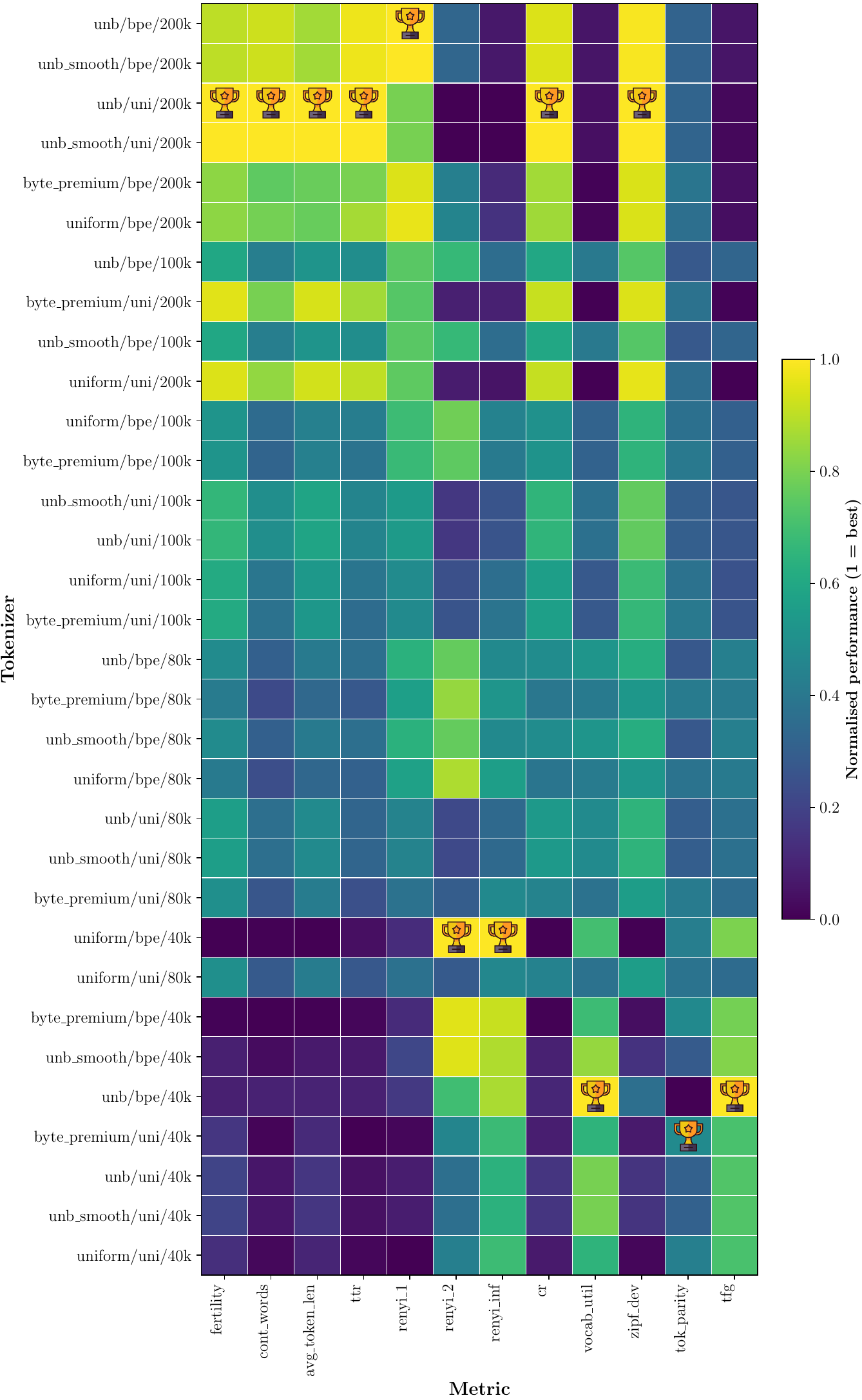}
    \caption{Text}
    \label{fig:text_tok_ranking}
  \end{subfigure}\hfill
  \begin{subfigure}[t]{0.49\textwidth}
    \centering
    \includegraphics[width=\linewidth]{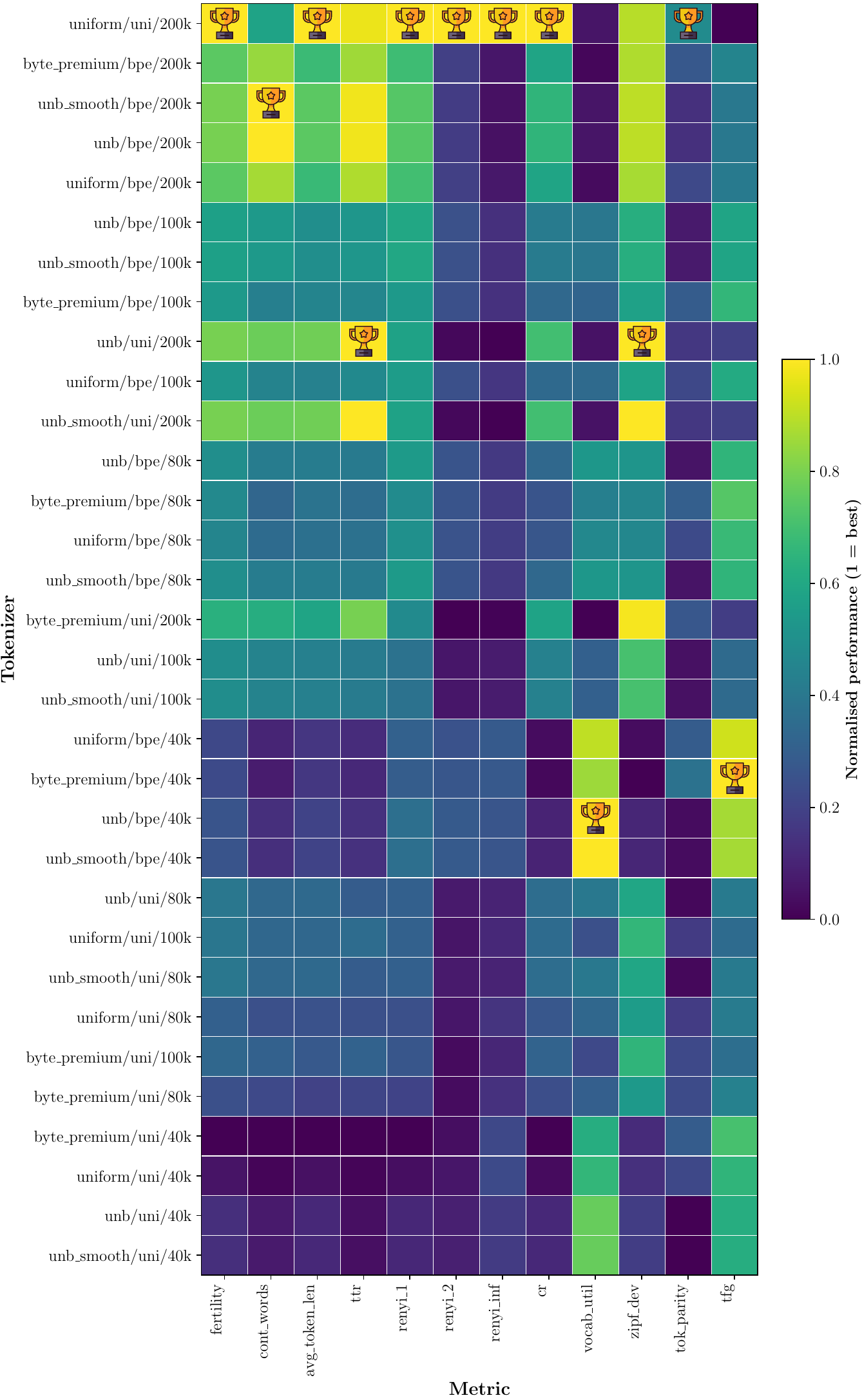}
    \caption{IPA}
    \label{fig:ipa_tok_ranking}
  \end{subfigure}
  \caption{Tokenizer selection. We rank Text and IPA tokenizers to select the optimal settings for GPT-2 pre-training. Optimal setting for Text is [BPE, 200k, data-proportional] and for IPA it is [UnigramLM, 200k, byte-uniform], according to the mean rank across metrics and languages.}
  \label{fig:tok_ranking}
\end{figure*}

\section{Tokenizer selection}\label{apdx:tok_selection}

Training autoregressive LMs from scratch for all tokenizer configurations is computationally prohibitive, so we use intrinsic evaluation to select a small set of tokenizers for GPT-2 pre-training. This appendix formalizes the ranking procedure and reports the full ranking heatmaps (Fig.~\ref{fig:tok_ranking}).

\paragraph{Set-up.}
Let $R\in\{\textsc{Text},\textsc{IPA}\}$ denote the input representation, and let $\mathcal{C}_R$ be the set of tokenizer configurations evaluated under $R$ (here $|\mathcal{C}_R|=32$). Each configuration $c\in\mathcal{C}_R$ is defined by the tuple
\[
c \;=\; (\text{algorithm}, \text{vocab size}, \text{sampling})\, .
\]
Let $\mathcal{L}$ be the set of evaluation languages and let $\mathcal{M}$ be the set of intrinsic metrics. For each language $\ell\in\mathcal{L}$, metric $m\in\mathcal{M}$, and configuration $c$, we obtain a per-language score $s_{m}(c,\ell)$ (as defined in Section~\ref{sec:intrinsic}).

\paragraph{Macro-averaged metric scores.}
To obtain a single score per metric and configuration, we macro-average over languages:
\begin{equation}
\bar{s}_{m}(c)
\;=\;
\frac{1}{|\mathcal{L}|}\sum_{\ell\in\mathcal{L}} s_{m}(c,\ell)\, .
\label{eq:macro_metric}
\end{equation}
In order to make the resulting heatmaps easier to read, we convert all metrics to a consistent higher-is-better orientation by defining:
\begin{equation}
\tilde{s}_{m}(c)=
\begin{cases}
\bar{s}_{m}(c) & \text{if } \uparrow,\\
-\bar{s}_{m}(c) & \text{if } \downarrow,\\
-\lvert \bar{s}_{m}(c)-1 \rvert & \text{if } \rightarrow1\leftarrow,
\end{cases}
\label{eq:metric_orientation}
\end{equation}
with arrows designating whether the original metric interpretation is higher-is-better ($\uparrow$), lower-is-better ($\downarrow$) or closer-to-one-is-better ($\rightarrow 1 \leftarrow$).

\begin{table*}[t]
\centering
\small
\setlength{\tabcolsep}{4pt}
\begin{tabular}{lcccccccccccc}
\toprule
& {\textbf{WF}$\downarrow$}
& {\textbf{PCW}$\downarrow$}
& {\textbf{ATL}$\uparrow$}
& {\textbf{TTR}$\uparrow$}
& {\textbf{RE\textsubscript{1}}$\uparrow$}
& {\textbf{RE\textsubscript{2}}$\uparrow$}
& {\textbf{RE\textsubscript{$\infty$}}$\uparrow$}
& {\textbf{CR}$\uparrow$}
& {\textbf{VU}$\uparrow$}
& {\textbf{ZipfD}$\downarrow$}
& {\textbf{TP}$\rightarrow 1 \leftarrow$}
& {\textbf{TFG}$\downarrow$} \\
\midrule
\textbf{\textcolor{text}{Text}} & 4.89 & 0.92 & 2.02 & 0.13 & 8.81 & \bfseries 6.37 & \bfseries 3.99 & 2.26 & 0.02 & 0.29 & 3.30 & 0.36 \\
\textbf{\textcolor{ipa}{IPA}}  & \bfseries 3.18 & \bfseries 0.90 & \bfseries 2.72 & \bfseries 0.16 & \bfseries 9.14 & 6.06 & 3.70 & \bfseries 2.98 & \bfseries 0.02 & \bfseries 0.16 & \bfseries 1.72 & \bfseries 0.17 \\
\bottomrule
\end{tabular}
\caption{Intrinsic tokenization quality at 300k vocabulary size, macro-averaged across languages. Arrows indicate whether lower ($\downarrow$), higher ($\uparrow$) or closer-to-1 ($\rightarrow 1 \leftarrow$) is better. Best values are boldfaced.}
\label{apdx:tab:300k}
\end{table*}

\paragraph{Per-metric ranks and mean rank aggregation.}
For each metric $m$ and representation $R$, we rank configurations by their oriented macro score $\tilde{s}_m(c)$:
\begin{equation}
r_{m}(c)
\;=\;
1 + \sum_{c'\in\mathcal{C}_R} \mathbb{I}\!\left[\tilde{s}_{m}(c') > \tilde{s}_{m}(c)\right];
\quad c\in\mathcal{C}_R,
\label{eq:rank_def}
\end{equation}
where $r_m(c)=1$ denotes the best configuration for metric $m$ (ties can be handled by average ranks; in practice ties are rare given continuous scores).
We aggregate metric-wise ranks into a single scalar score by averaging across metrics:
\begin{equation}
\mathrm{MR}(c)
\;=\;
\frac{1}{|\mathcal{M}|}\sum_{m\in\mathcal{M}} r_{m}(c)\, .
\label{eq:mean_rank}
\end{equation}
Lower $\mathrm{MR}(c)$ indicates a better overall intrinsic ranking. The full rank matrices $\{r_m(c)\}$ and mean-rank scores $\mathrm{MR}(c)$ are visualized in Fig.~\ref{fig:text_tok_ranking} (\textcolor{text}{\textbf{Text}}) and Fig.~\ref{fig:ipa_tok_ranking} (\textcolor{ipa}{\textbf{IPA}}).

\paragraph{Selecting Opt and Subopt tokenizers.}
For each representation $R$, we select the intrinsically best tokenizer as
\begin{equation}
c_R^{\star} \;=\; \arg\min_{c\in\mathcal{C}_R}\mathrm{MR}(c)\, .
\label{eq:opt_select}
\end{equation}
We refer to the resulting tokenizers as \textcolor{text}{\emph{\textbf{Text Opt}}} and \textcolor{ipa}{\emph{\textbf{IPA Opt}}}. In our experiments, \textcolor{text}{\emph{\textbf{Text Opt}}} corresponds to the configuration $c_{\text{Text}}^{\star} = $ \texttt{(BPE, 200k, data-proportional)}, whereas \textcolor{ipa}{\emph{\textbf{IPA Opt}}} corresponds to $c_{\text{IPA}}^{\star} = $ \texttt{(UnigramLM, 200k, byte-uniform)}.
Because these optima arise under different hyperparameter settings, they differ along both representation and tokenizer configuration axes. To control for this, we also select cross-swapped ``Subopt'' tokenizers: a Text tokenizer trained with the IPA Opt configuration (\textcolor{textsub}{\emph{\textbf{Text Subopt}}}) and an IPA tokenizer trained with the Text Opt configuration (\textcolor{ipasub}{\emph{\textbf{IPA Subopt}}}). We then pre-train four GPT-2 models in total: Text Opt, Text Subopt, IPA Opt, and IPA Subopt.

\paragraph{Larger vocabulary sizes.} As the optimal tokenizers for both \textbf{\textcolor{text}{Text}} and \textbf{\textcolor{ipa}{IPA}} are found on the largest vocabulary size we tested (200k), we consider an even larger vocabulary size of 300k. We find that IPA still outperforms Text on intrinsic metrics with similar trends, though IPA tokenizer's intrinsic quality slightly degrades relative to its 200k version (while TFG improves, suggesting better cross-language balance). Text tokenizer quality is similar between 200k and 300k. However, the previously found optimal settings for both input representations remain optimal even when considering the 300k vocabulary. We provide the results in Table~\ref{apdx:tab:300k}.

\section{Experimental setup}\label{apdx:exp_setup}

\subsection{Tokenizer training hyper-parameters}\label{apdx:tok_HPs}

We provide tokenizer training hyper-parameters in Table~\ref{tab:tokenizer_configurations}.

\begin{table*}[t]
\small
\centering
\begin{tabular}{@{}lll@{}}
\toprule
\textbf{Mode} & \textbf{Parameter} & \textbf{Value} \\
\midrule
\multirow{6}{*}{Text} 
  & \texttt{model\_type}         & [bpe, unigram] \\
  & \texttt{vocab\_size}         & [40\,000, 80\,000, 100\,000, 200\,000] \\
  & \texttt{character\_coverage} & 0.9995 \\
  & \texttt{split\_by\_unicode\_script} & True \\
  & \texttt{byte\_fallback} & True \\
  & \texttt{normalization\_rule\_name} & nmt\_nfkc \\
\midrule
\multirow{6}{*}{IPA} 
  & \texttt{model\_type}         & [bpe, unigram] \\
  & \texttt{vocab\_size}         & [40\,000, 80\,000, 100\,000, 200\,000] \\
  & \texttt{character\_coverage} & 1.0 \\
  & \texttt{split\_by\_unicode\_script} & False \\
  & \texttt{byte\_fallback} & False \\
  & \texttt{normalization\_rule\_name} & identity \\
\bottomrule
\end{tabular}
\caption{Tokenizer training hyper-parameter settings used for Text and IPA representations. We show only the relevant parameters, for all others we used the default SentencePiece values for both modes. Values enclosed in square brackets signify that multiple configurations were considered.}
\label{tab:tokenizer_configurations}
\end{table*}

\subsection{GPT-2 training and fine-tuning hyper-parameters}\label{apdx:gpt2_HPs}

We provide GPT-2 pre-training hyper-parameters in Table~\ref{tab:gpt2_hparams} and fine-tuning hyper-parameters in Table~\ref{tab:ft_hparams}.

\begin{table}[t]
\centering
\small
\begin{tabular}{@{}ll@{}}
\toprule
\textbf{Hyperparameter} & \textbf{Value} \\
\midrule
Architecture (layers/heads/hidden) & 12 / 12 / 768 \\
Context length & 2048 \\
Vocabulary size & 200k \\
Training steps & 10,000 \\
Batch size (effective) & 64 \\
Learning rate & $5 \times 10^{-4}$ \\
Scheduler & cosine \\
Warmup ratio & 0.03 \\
Weight decay & 0.1 \\
Precision & bfloat16 \\
Attention implementation & SDPA \\
Random seed & 42 \\
\bottomrule
\end{tabular}
\caption{GPT-2 pretraining hyperparameters, shared across all Text/IPA and Opt/Subopt conditions.}
\label{tab:gpt2_hparams}
\end{table}

\begin{table}[t]
\centering
\small
\begin{tabular}{@{}lcc@{}}
\toprule
\textbf{Hyperparameter} & \textbf{XNLI} & \textbf{PAWS-X} \\
\midrule
Number of labels & 3 & 2 \\
Training steps & 6128 & 3860 \\
Batch size (effective) & 256 & 256 \\
Learning rate & $1 \times 10^{-4}$ & $3 \times 10^{-5}$ \\
Scheduler & cosine & cosine \\
Warmup ratio & 0.06 & 0.06 \\
Weight decay & 0.0 & 0.0 \\
Max grad norm & 1.0 & 1.0 \\
Precision & bfloat16 & bfloat16 \\
Model selection & validation loss & validation loss \\
\bottomrule
\end{tabular}
\caption{Fine-tuning hyperparameters for XNLI and PAWS-X, shared across all Text/IPA and Opt/Subopt conditions within each task.}
\label{tab:ft_hparams}
\end{table}

\subsection{Computational footprint}\label{apdx:comp_footprint}

All experiments were run on a single GPU on the Snellius cluster, using either an NVIDIA H100 GPU (for GPT-2 pre-training runs) or an NVIDIA A100 GPU (for all other experiments). The runs conducted on a single A100 GPU used 85.74 GPU-hours, corresponding to an estimated 21.43 kWh of electricity consumption and 5.61 kg CO2eq. The runs conducted on a single H100 GPU used 27.93 GPU-hours, corresponding to an estimated 19.55 kWh and 5.12 kg CO2eq. In total, our experiments used 113.67 GPU hours, 40.98 kWh of electricity, and produced 10.73 kg CO2eq. Estimations were conducted using the \href{https://mlco2.github.io/impact#compute}{MachineLearning Impact calculator} presented in \citet{lacoste2019quantifying}.

\section{General benefits of IPA as an input representation}\label{apdx:ipa_benefits}

In the introduction section, we have outlined several beneficial properties offered by the IPA. Here we show what implications those properties have in practice.

\begin{figure*}[t]
  \centering
  \includegraphics[width=1.5\columnwidth]{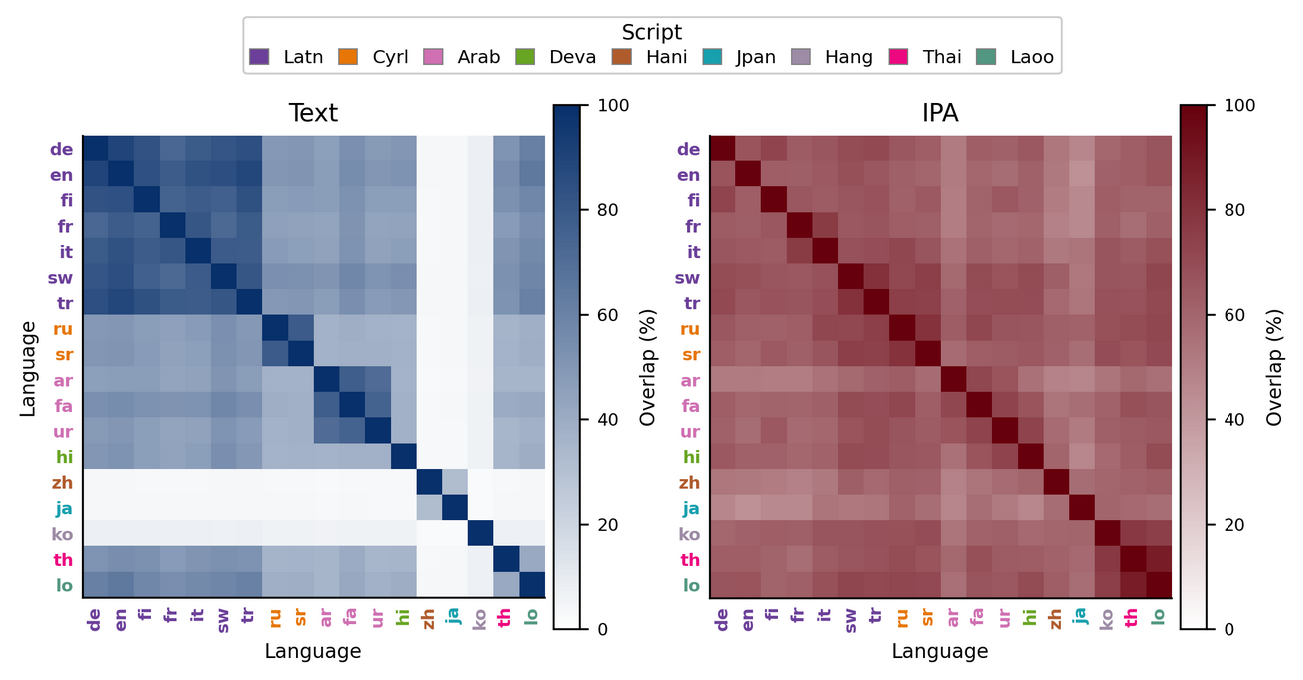}
  \caption{Pairwise overlap of character inventories (\%) between the 18 pre-training languages on the CulturaX \texttt{byte-uniform} sample, computed on the original \textcolor{text}{\textbf{Text}} input (left) and its \textcolor{ipa}{\textbf{IPA}} transcription (right). Each cell reports the percentage of unique characters shared by a language pair (diagonal = 100\%). In \textcolor{text}{\textbf{Text}}, overlap is largely confined to languages that share a script, with near-zero sharing for languages with distinct orthographies (e.g., \textsc{zh}, \textsc{ja}, \textsc{ko}); in \textcolor{ipa}{\textbf{IPA}}, overlap becomes higher across languages and more uniform across scripts due to the shared phonemic symbol inventory.}
  \label{fig:char_sharing}
\end{figure*}

\begin{figure}[t]
  \centering
  \includegraphics[width=.75\columnwidth]{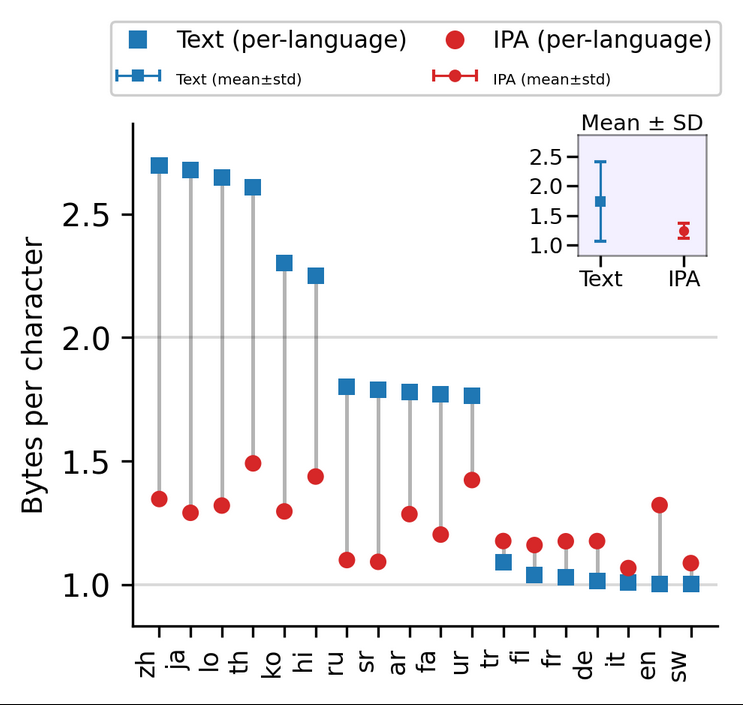}
  \caption{Average bytes-per-character (BPC) by language on the CulturaX \texttt{byte-uniform} sample, for the original \textcolor{text}{\textbf{Text}} input and its \textcolor{ipa}{\textbf{IPA}} transcription. \textcolor{text}{\textbf{Text}} exhibits large cross-lingual disparities, with higher BPC for languages whose scripts occupy higher Unicode ranges (e.g., \textsc{zh}, \textsc{ja}, \textsc{lo}, \textsc{th}), while Latin-script languages cluster near 1 BPC. \textcolor{ipa}{\textbf{IPA}} concentrates most symbols in low-byte code points, reducing mean BPC and making storage cost more uniform across languages; the inset (top-right) summarizes the mean $\pm$ standard deviation across languages.}
  \label{fig:bpc}
\end{figure}

\paragraph{IPA improves cross-lingual sharing of character inventories.} In Figure~\ref{fig:char_sharing} we show the percentage of shared characters across each pair of our 18 languages on the CulturaX sample we used for pre-training on the original text and on the converted IPA text. In Text heatmap, we see biggest sharing between languages that use the same script, but that percentage drops for language pairs with different scripts. We see that languages with complex orthographies (Chinese, Japanese, and Korean) have almost zero overlap with any other language. Other languages do show some percentage of overlap even though they differ in scripts, partly because many languages in our sample do contain inputs from other languages, most notably English. However, the IPA heatmap reveals a much more evenly distributed and overall larger character overlap across languages. This is a direct consequence of IPA's shared input representation for all of the languages. Overall, this increased sharing is beneficial for multilingual tokenizers, as cross-lingual token overlap has been shown to improve model performance~\cite{zhang-etal-2025-tomato, kallini-etal-2025-false}.

\paragraph{IPA reduces storage overhead across languages on average.} Figure~\ref{fig:bpc} compares the average number of bytes needed to encode a single character per language for Text vs IPA, estimated from our pre-training CulturaX sample. Standard text representation shows notable disparities: languages with complex orthographies (Chinese, Japanese, Lao, and Thai) require much larger number ($>2.5$) of bytes-per-character (BPC), while languages using Latin script are all very close to $1$ byte-per-character. This difference is due to the position of different characters in the Unicode range. The symbols used by the IPA mostly use ASCII characters (1 BPC), together with a set of special IPA symbols (2 BPC). Some additional, but less frequent IPA markers require 3 BPC. This fact, together with IPA's shared language representation space, reduces the average BPC across our set of languages, requiring slightly larger numbers for Latin-script languages than Text, but much lower BPC for all other ones. In addition, the required storage memory is much more balanced across languages.

\section{IPA for byte-level tokenization}\label{apdx:ipa_byte_level}

\begin{table}[!t]
\centering
\small
\setlength{\tabcolsep}{7pt}
\begin{tabular}{c c c}
\toprule
\textbf{Lang.} & \textbf{\textcolor{text}{Text}} & \textbf{\textcolor{ipa}{IPA}} \\
\midrule
\textsc{am} & 220 & \textbf{198} \\
\textsc{ar} & 204 & \textbf{188} \\
\textsc{de} & \textbf{151} & 186 \\
\textsc{el} & 278 & \textbf{153} \\
\textsc{en} & \textbf{128} & 161 \\
\textsc{es} & \textbf{155} & 160 \\
\textsc{fa} & 218 & \textbf{146} \\
\textsc{fi} & \textbf{141} & 160 \\
\textsc{fr} & \textbf{158} & 169 \\
\textsc{he} & 177 & \textbf{118} \\
\textsc{hi} & 328 & \textbf{219} \\
\textsc{it} & \textbf{152} & 162 \\
\textsc{ja} & 162 & \textbf{155} \\
\textsc{ko} & \textbf{153} & 173 \\
\textsc{lo} & 351 & \textbf{207} \\
\textsc{my} & 452 & \textbf{251} \\
\textsc{pl} & \textbf{144} & 174 \\
\textsc{ru} & 255 & \textbf{183} \\
\textsc{sr} & 231 & \textbf{141} \\
\textsc{sw} & \textbf{134} & 146 \\
\textsc{th} & 354 & \textbf{218} \\
\textsc{tr} & \textbf{143} & 164 \\
\textsc{ur} & \textbf{225} & 246 \\
\textsc{zh} & \textbf{118} & 167 \\
\midrule
\textit{mean\ $\pm$ std.} & 210 $\pm$ 87 & \textbf{177 $\pm$ 32} \\
\bottomrule
\end{tabular}
\caption{Average number of tokens per sentence on FLORES+ under byte-level tokenization for Text and IPA input. Boldface marks the lower value within each row (lower values are better). Macro-level mean$\pm$std. are shown at the bottom.}
\label{apdx:tab:ipa_byte_level}
\end{table}

The benefits of IPA as an input representation highlighted in Appendix~\ref{apdx:ipa_benefits} also translate to byte-level tokenization-free approaches. In Table~\ref{apdx:tab:ipa_byte_level} we empirically show that, when considering individual bytes as tokens, working with IPA input still results in shorter average sequence lengths compared to using Text inputs. It also reduces the standard deviation, making the sequence length distribution more stable across languages. Therefore, we believe IPA has a potential to be applied on top of the existing byte-level methods to further increase cross-script robustness.

\section{Per-language results for suboptimal tokenizers}\label{apdx:subopt_per_lang}

\begin{figure*}[t]
  \centering
  \includegraphics[width=2\columnwidth]{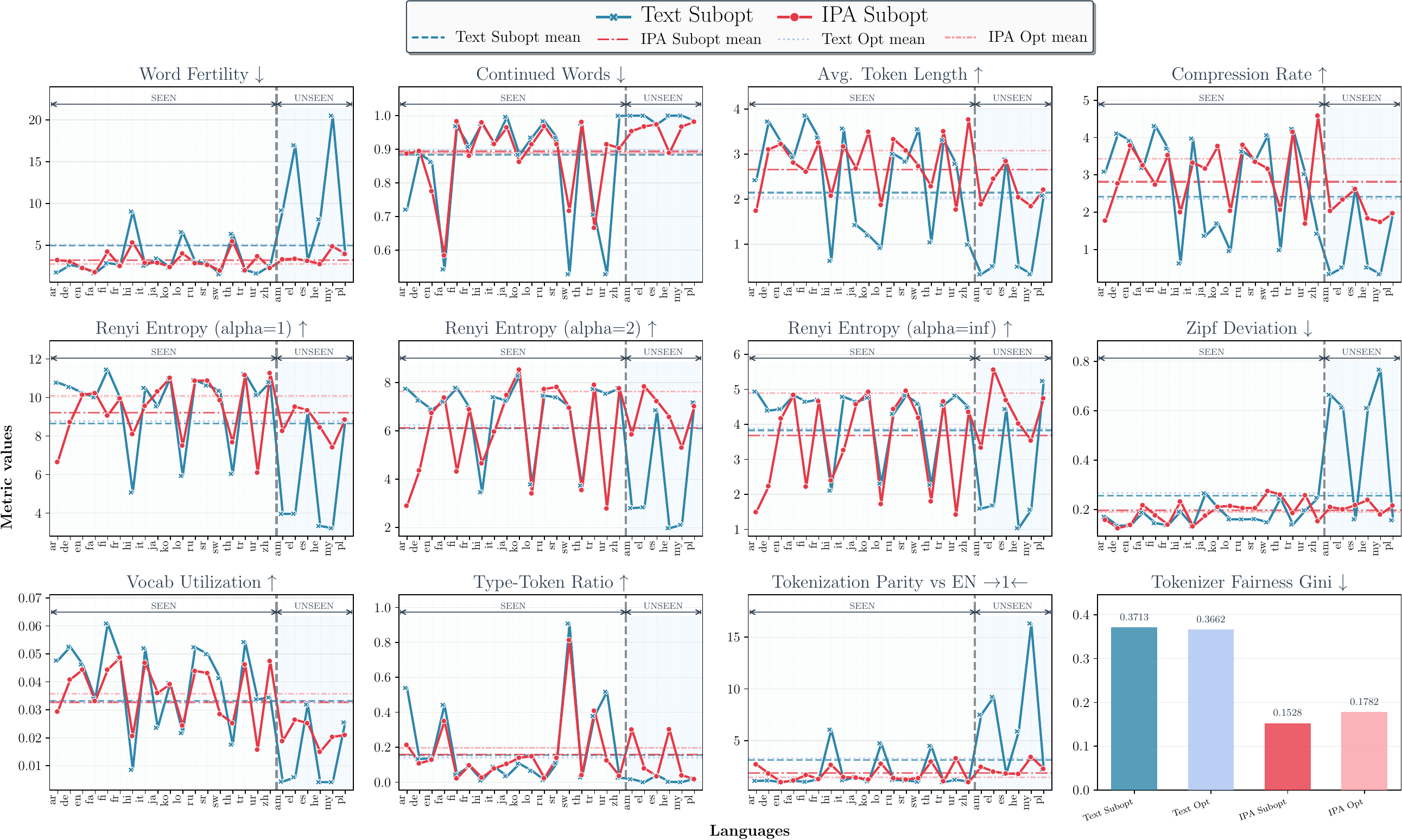}
  \caption{Detailed per-language performance of the suboptimal tokenizers.}
  \label{fig:subopt_per_lang}
\end{figure*}

We provide a detailed per-language intrinsic evaluation of the suboptimal tokenizers in Fig.~\ref{fig:subopt_per_lang}.

\begin{figure*}[t]
  \centering
  \includegraphics[width=2\columnwidth]{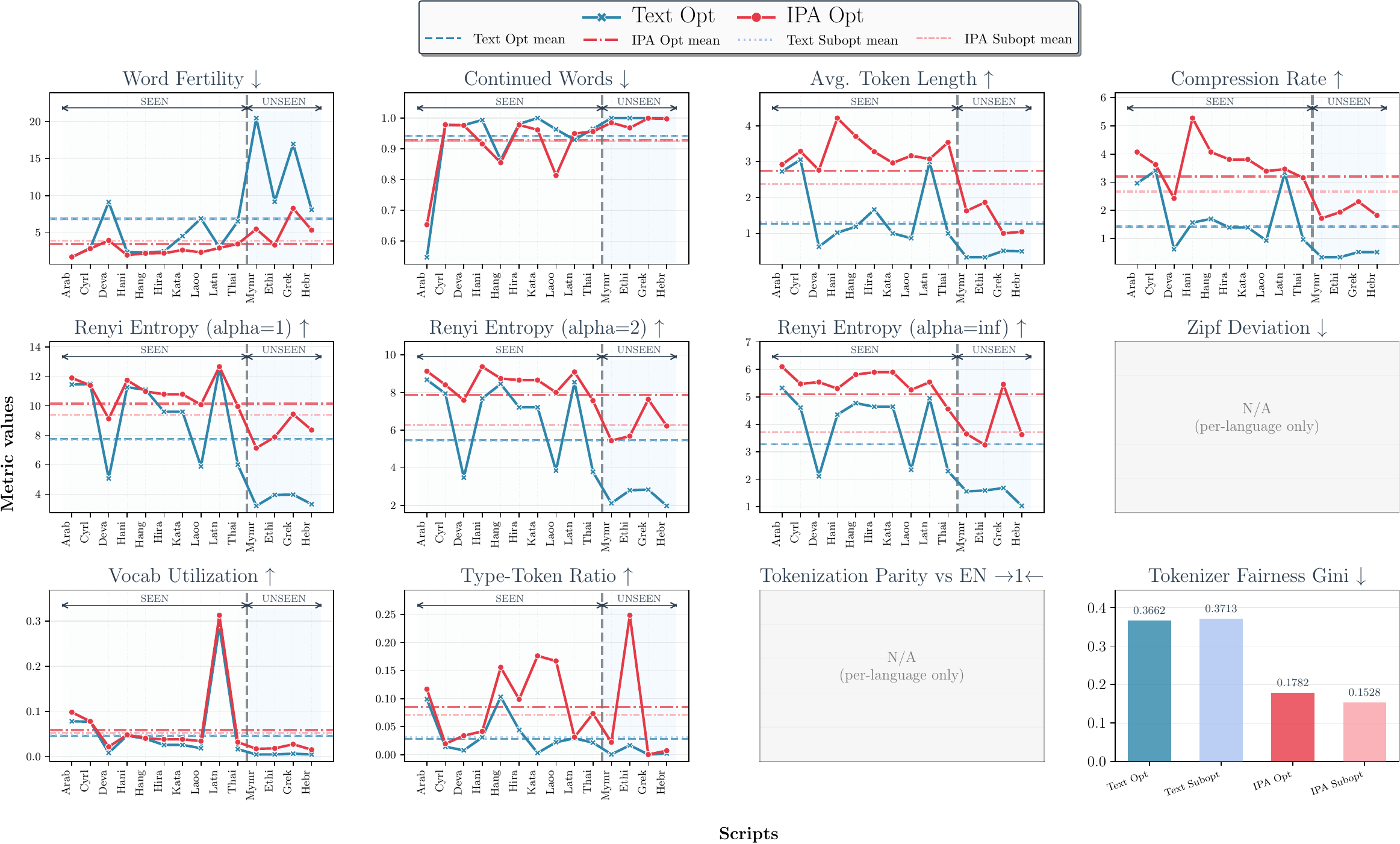}
  \caption{Per-script intrinsic tokenization metrics: \textcolor{text}{\textbf{Text Opt}} vs.\ \textcolor{ipa}{\textbf{IPA Opt}}. Solid lines show \emph{Text Opt} and \emph{IPA Opt} per language; dashed lines are means. Arrows: $\uparrow$ = higher is better, $\downarrow$ = lower is better, $\rightarrow1\leftarrow$ = closer-to-1 is better; TFG is a single global tokenizer metric (not per-language), hence shown as bars. Zipf Deviation and Tokenization Parity are per-language only, so they are omitted here. \textsc{Seen}/\textsc{Unseen} mark languages included in vs.\ absent from tokenizer training data. \textbf{Takeaway:} \emph{IPA} improves tokenization quality for non-Latin scripts, while maintaining the same quality on Latin script languages as \emph{Text} tokenizers. Improvements are most drastic for scripts unseen during tokenizer training.}
  \label{fig:intrinsic_per_script_opt}
\end{figure*}

\section{Detailed per-script diagnostic}\label{apdx:per_script_details}

\begin{figure*}[t]
  \centering
  \includegraphics[width=2\columnwidth]{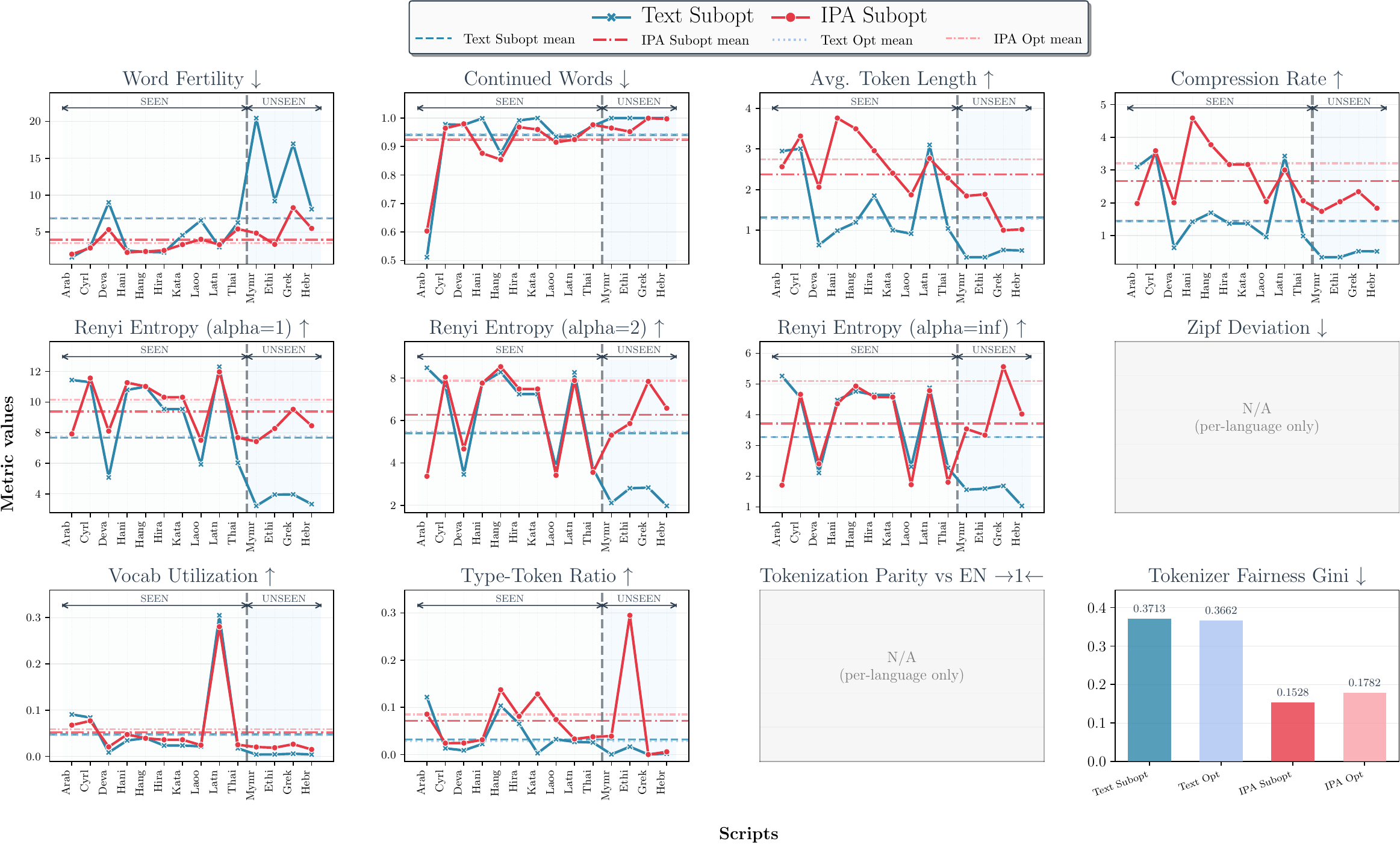}
  \caption{Per-script intrinsic tokenization metrics: \textcolor{textsub}{\textbf{Text Subopt}} vs.\ \textcolor{ipasub}{\textbf{IPA Subopt}}. Solid lines show \emph{Text Subopt} and \emph{IPA Subopt} per language; dashed lines are means. Arrows: $\uparrow$ = higher is better, $\downarrow$ = lower is better, $\rightarrow1\leftarrow$ = closer-to-1 is better; TFG is a single global tokenizer metric (not per-language), hence shown as bars. Zipf Deviation and Tokenization Parity are per-language only, so they are omitted here. \textsc{Seen}/\textsc{Unseen} mark languages included in vs.\ absent from tokenizer training data.}
  \label{fig:intrinsic_per_script_subopt}
\end{figure*}

We provide a detailed per-script intrinsic evaluation for both the optimal (Fig.~\ref{fig:intrinsic_per_script_opt}) and suboptimal tokenizers (Fig.~\ref{fig:intrinsic_per_script_subopt}).

\section{Interpretation of tokenizer configuration effects}\label{apdx:config_effects}

\paragraph{Compression metrics improve with vocab size.} This is consistent across algorithms, sampling strategies, and modes (text/ipa). Consequently, vocab size of 200k performs best in terms of compression (WF, PCW, ATL, CR). This is expected, as larger vocabs will have more space for storing tokens, which means fewer tokens needed to represent the same input.

\paragraph{Renyi $H_2$ and $H_\infty$ best at smallest vocab; $H_1$ best at 200k.} This pattern is consistent with the different sensitivities of R'enyi orders: $H_1$ (Shannon entropy) reflects the full distribution and increases when the vocabulary grows to include many rare, low-probability types (e.g., longer tokens approaching full words). In contrast, $H_2$ and especially $H_\infty$ place increasing weight on the head of the distribution and thus tend to favor settings where probability mass is less concentrated in a small set of very frequent tokens—often achieved by smaller vocabularies that encourage greater sharing of subword units. Because the entropy scale depends strongly on vocabulary size, we treat these trends primarily as descriptive of configuration effects and focus comparisons on the Text vs.\ IPA axis under matched vocabulary sizes. One exception is \texttt{[IPA, UnigramLM, byte-uniform]}, where all three entropies are maximized at 200k; we leave a more detailed analysis of this case to future work.

\paragraph{Zipf Deviation decreases with vocabulary size.} Across all settings, Zipf Deviation (ZipfD) is lowest at the largest vocabulary (200k). In Fig.~\ref{fig:heatmap}, ZipfD appears as higher-is-better because we invert and normalize metrics for readability; in the original definition, lower ZipfD indicates that the empirical rank--frequency curve is closer to an ideal Zipfian trend. The observed monotonic improvement with vocabulary size is intuitive: as vocabularies grow, token inventories increasingly contain longer units (often close to whole-word tokens), and the resulting token frequency distribution more closely mirrors the underlying word-frequency distribution, which is approximately Zipfian in natural language. At the same time, ZipfD is strongly affected by vocabulary granularity, so we interpret the vocabulary-size trend as a configuration effect and emphasize comparisons along the Text vs.\ IPA axis under matched vocabulary sizes.

\paragraph{Vocabulary utilization always best for 40k, while TTR always best for 200k.} Although both metrics quantify vocabulary usage, they capture different aspects of the token distribution. VU measures the fraction of the learned vocabulary that is observed at least once on evaluation data; smaller vocabularies encourage heavier reuse of a limited set of subword units, making it more likely that most entries are encountered. In contrast, TTR measures the number of distinct token types relative to the total number of produced tokens; larger vocabularies introduce more specific (and often rarer) token types, increasing the diversity of observed types even if many are used only infrequently.

\section{Additional statistics: win-rate, paired effect size, and mean z-score}\label{app:wr_dz_zbar}

We summarize intrinsic differences between \textsc{IPA} and \textsc{Text} using (i) a per-language win-rate, (ii) a paired effect size $d_z$ per metric, and (iii) an overall per-language score based on mean $z$-scored improvements. All computations are paired on the intersection of languages (or scripts) available for both representations. We exclude TFG from per-language/script computations because it is a global metric.

\paragraph{Direction-corrected improvement.}
For metric $m$ and language/script $i$, let $x^{\textsc{IPA}}_{m,i}$ and $x^{\textsc{Text}}_{m,i}$ be the corresponding metric values. We convert each metric to an ``IPA-better'' improvement value $\Delta_{m,i}$:
\begin{equation}
\Delta_{m,i}=
\begin{cases}
x^{\textsc{IPA}}_{m,i}-x^{\textsc{Text}}_{m,i} & (\uparrow)\\
x^{\textsc{Text}}_{m,i}-x^{\textsc{IPA}}_{m,i} & (\downarrow)\\
|x^{\textsc{Text}}_{m,i}-1|-|x^{\textsc{IPA}}_{m,i}-1| & (\rightarrow 1\leftarrow)
\end{cases}
\label{eq:delta_simple}
\end{equation}
Here $(\uparrow)$ denotes metrics where larger values are better, $(\downarrow)$ denotes metrics where smaller values are better, and $(\rightarrow 1 \leftarrow)$ denotes metrics where values closer to $1$ are better. Thus, $\Delta_{m,i}>0$ means \textsc{IPA} is better than \textsc{Text} on metric $m$ for item $i$.

\paragraph{Win-rate (per language/script).}
Let $\mathcal{M}_i$ be the set of metrics available for item $i$. The win-rate is the fraction of metrics improved by \textsc{IPA}:
\begin{equation}
\mathrm{WinRate}(i)=\frac{1}{|\mathcal{M}_i|}\sum_{m\in\mathcal{M}_i}\mathbb{I}\!\left[\Delta_{m,i}>0\right].
\label{eq:winrate_simple}
\end{equation}

\paragraph{Paired effect size $d_z$ (per metric).}
For each metric $m$, we compute the mean and standard deviation of $\Delta_{m,i}$ across items $i\in\mathcal{I}_m$ (the items with non-missing paired values):
\begin{equation}
\begin{aligned}
\bar{\Delta}_m
&=
\frac{1}{|\mathcal{I}_m|}
\sum_{i\in\mathcal{I}_m}\Delta_{m,i}, \\
s_m
&=
\mathrm{std}\!\left(\{\Delta_{m,i}\}_{i\in\mathcal{I}_m}\right).
\end{aligned}
\label{eq:dz_stats_simple}
\end{equation}
We then report Cohen's paired standardized mean difference:
\begin{equation}
d_z(m)=\frac{\bar{\Delta}_m}{s_m+\varepsilon}.
\label{eq:dz_simple}
\end{equation}
where $\varepsilon$ is a small constant for numerical stability. Positive $d_z$ indicates that \textsc{IPA} tends to outperform \textsc{Text} on metric $m$.

\paragraph{Mean $z$-score (overall per language/script).}
To combine metrics with different scales, we $z$-score improvements within each metric across items:
\begin{equation}
z_{m,i}=\frac{\Delta_{m,i}-\mu_m}{\sigma_m},
\label{eq:z_simple}
\end{equation}
where $\mu_m$ and $\sigma_m$ are the mean and standard deviation of $\Delta_{m,i}$ over $i\in\mathcal{I}_m$ (if $\sigma_m=0$, we set $z_{m,i}=0$). The overall score for item $i$ is the mean $z$ across its available metrics:
\begin{equation}
\mathrm{MeanZ}(i)=\frac{1}{|\mathcal{M}_i|}\sum_{m\in\mathcal{M}_i} z_{m,i}.
\label{eq:meanz_simple}
\end{equation}
Higher $\mathrm{MeanZ}(i)$ indicates that \textsc{IPA} improves more metrics, and by a larger margin, relative to \textsc{Text}.

\section{Sensitivity to noise injection}\label{app:noise}

As an additional experiment, we test robustness of our \textcolor{ipa}{\textbf{IPA Opt}} tokenizer to noise injection and compare it to that of the \textcolor{text}{\textbf{Text Opt}} tokenizer. Specifically, we test sensitivity to random character deletions with three different rates (5\%, 10\%, and 15\%). We provide results in Table~\ref{tab:noise}.

\begin{table*}[t]
\centering
\small
\setlength{\tabcolsep}{5pt}
\renewcommand{\arraystretch}{1.15}
\resizebox{\textwidth}{!}{%
\begin{tabular}{clccccccccccc}
\toprule
\textbf{Mode} & \textbf{Del. rate} & \textbf{WF$\downarrow$} & \textbf{PCW$\downarrow$} & \textbf{ATL$\uparrow$} & \textbf{TTR$\uparrow$} & \textbf{RE\textsubscript{1}$\uparrow$} & \textbf{RE\textsubscript{2}$\uparrow$} & \textbf{RE\textsubscript{$\infty$}$\uparrow$} & \textbf{CR$\uparrow$} & \textbf{VU$\uparrow$} & \textbf{ZipfD$\downarrow$} & \textbf{TP$\rightarrow 1 \leftarrow$} \\
\midrule
\multirow{3}{*}{\rotatebox[origin=c]{90}{\textcolor{text}{\textbf{Text Opt}}}}
  & 5\%  & 5.89 & 0.90 & 1.99 & 0.13 & 9.39 & 6.70 & 4.17 & 2.47 & 0.04 & 0.24 & 2.67 \\
  & 10\% & 6.02 & 0.90 & 1.90 & 0.12 & 9.28 & 6.58 & 4.08 & 2.41 & 0.04 & 0.25 & 2.74 \\
  & 15\% & 6.15 & 0.90 & 1.83 & 0.12 & 9.16 & 6.46 & 3.99 & 2.35 & 0.03 & 0.26 & 2.82 \\
\midrule
\multirow{3}{*}{\rotatebox[origin=c]{90}{\textcolor{ipa}{\textbf{IPA Opt}}}}
  & 5\%  & 2.84 & 0.90 & 2.85 & 0.17 & 10.20 & 7.75 & 4.97 & 3.52 & 0.04 & 0.19 & 1.42 \\
  & 10\% & 2.85 & 0.91 & 2.66 & 0.16 & 10.08 & 7.63 & 4.88 & 3.46 & 0.04 & 0.20 & 1.47 \\
  & 15\% & 2.83 & 0.91 & 2.52 & 0.16 & 9.95 & 7.50 & 4.79 & 3.39 & 0.03 & 0.21 & 1.53 \\
\bottomrule
\end{tabular}%
}
\caption{Sensitivity to noise injection (random character deletions) of \textcolor{text}{\textbf{Text Opt}} and \textcolor{ipa}{\textbf{IPA Opt}} tokenizers.}
\label{tab:noise}
\end{table*}

\end{document}